\documentclass[nonacm,sigconf]{acmart}

\AtBeginDocument{%
  }

\usepackage{enumitem}
\usepackage{tabularx}
\usepackage{booktabs}

\setcopyright{none}

\copyrightyear{2025}
\begin{document}

\title{In the Blink of an Eye: Instant Game Map Editing using a Generative-AI Smart Brush}

\author{Vitaly Gnatyuk}
\affiliation{%
  \institution{Wargaming}
  \city{Berlin}
  \country{Germany}}
\email{v_gnatyuk@wargaming.net}

\author{Valeriia Koriukina}
\affiliation{%
  \institution{Wargaming}
  \city{Nicosia}
  \country{Cyprus}}
\email{v_koriukina@wargaming.net}

\author{Ilya Levoshevich}
\affiliation{%
  \institution{Wargaming}
  \city{Nicosia}
  \country{Cyprus}}
\email{i_levoshevich@wargaming.net}

\author{Pavel Nurminskiy}
\affiliation{%
  \institution{Wargaming}
  \city{Nicosia}
  \country{Cyprus}}
\email{p_nurminskiy@wargaming.net}

\author{G\"unter Wallner}
\affiliation{%
  \institution{Johannes Kepler University Linz}
  \city{Linz}
  \country{Austria}}
\email{guenter.wallner@jku.at}

\renewcommand{\shortauthors}{Gnatyuk et al.}

\begin{abstract}
    With video games steadily increasing in complexity, automated generation of game content has found widespread interest. However, the task of 3D gaming map art creation remains underexplored to date due to its unique complexity and domain-specific challenges. While recent works have addressed related topics such as retro-style level generation and procedural terrain creation, these works primarily focus on simpler data distributions. To the best of our knowledge, we are the first to demonstrate the application of modern AI techniques for high-resolution texture manipulation in complex, highly detailed AAA 3D game environments. We introduce a novel \emph{Smart Brush} for map editing, designed to assist artists in seamlessly modifying selected areas of a game map with minimal effort. By leveraging generative adversarial networks and diffusion models we propose two variants of the brush that enable efficient and context-aware generation. Our hybrid workflow aims to enhance both artistic flexibility and production efficiency, enabling the refinement of environments without manually reworking every detail, thus helping to bridge the gap between automation and creative control in game development. A comparative evaluation of our two methods with adapted versions of several state-of-the art models shows that our GAN-based brush produces the sharpest and most detailed outputs while preserving image context while the evaluated state-of-the-art models tend towards blurrier results and exhibit difficulties in maintaining contextual consistency.
\end{abstract}




\begin{teaserfigure}
    \centering
    \includegraphics[width=0.9\textwidth]{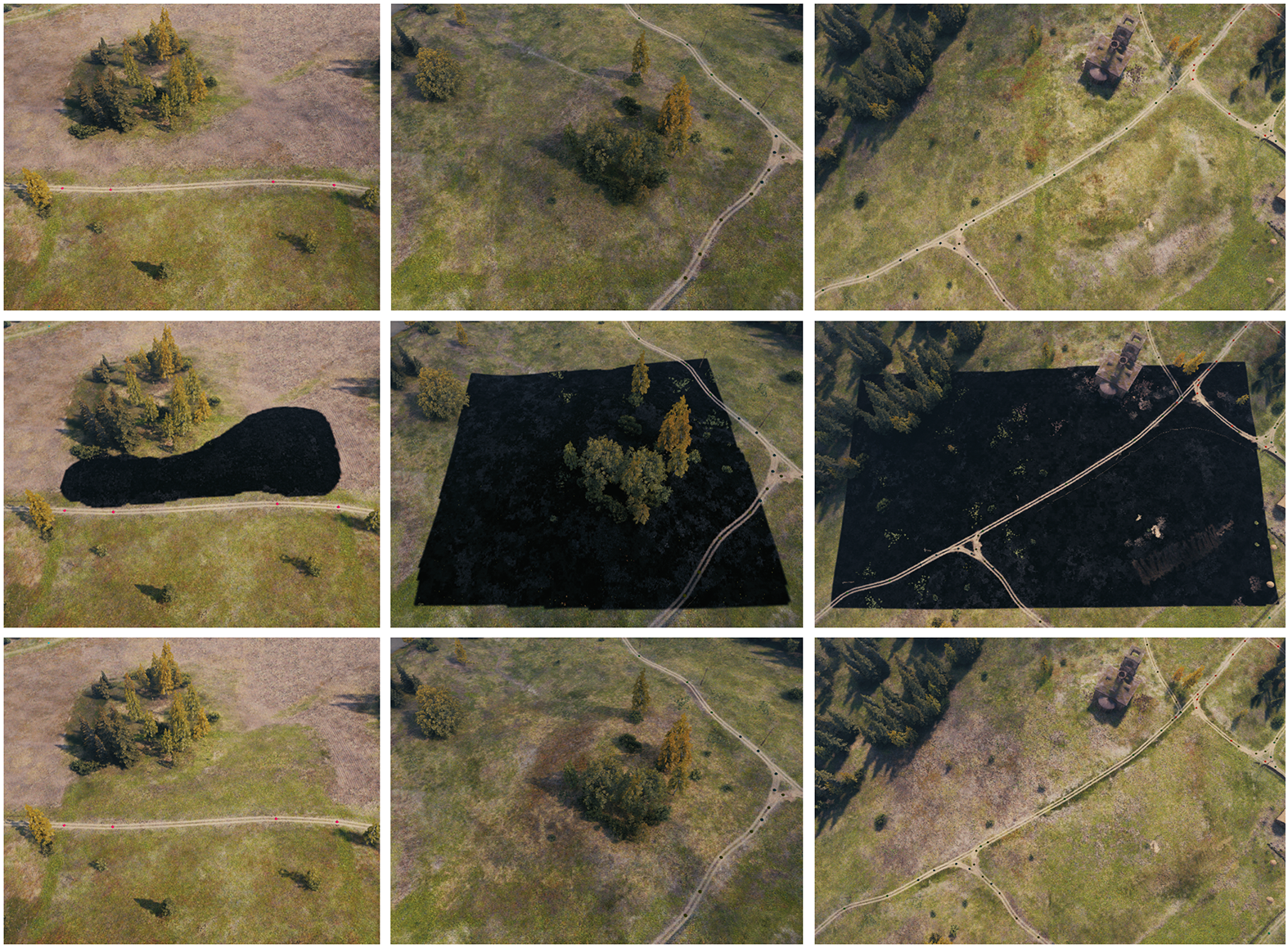}
    \caption{Examples of applying our \emph{Smart Brush}: The top row shows chunks from the original map, the second row shows the user-defined mask (blackish areas) within the generation should be performed, and the bottom row displays the chunks generated using our \emph{Smart Brush} approach. }
    \label{fig:teaser}
\end{teaserfigure}

\maketitle

\section{Introduction}
    \label{sec:introduction}

\begin{figure}
    \centering
    \includegraphics[width=1.0\linewidth]{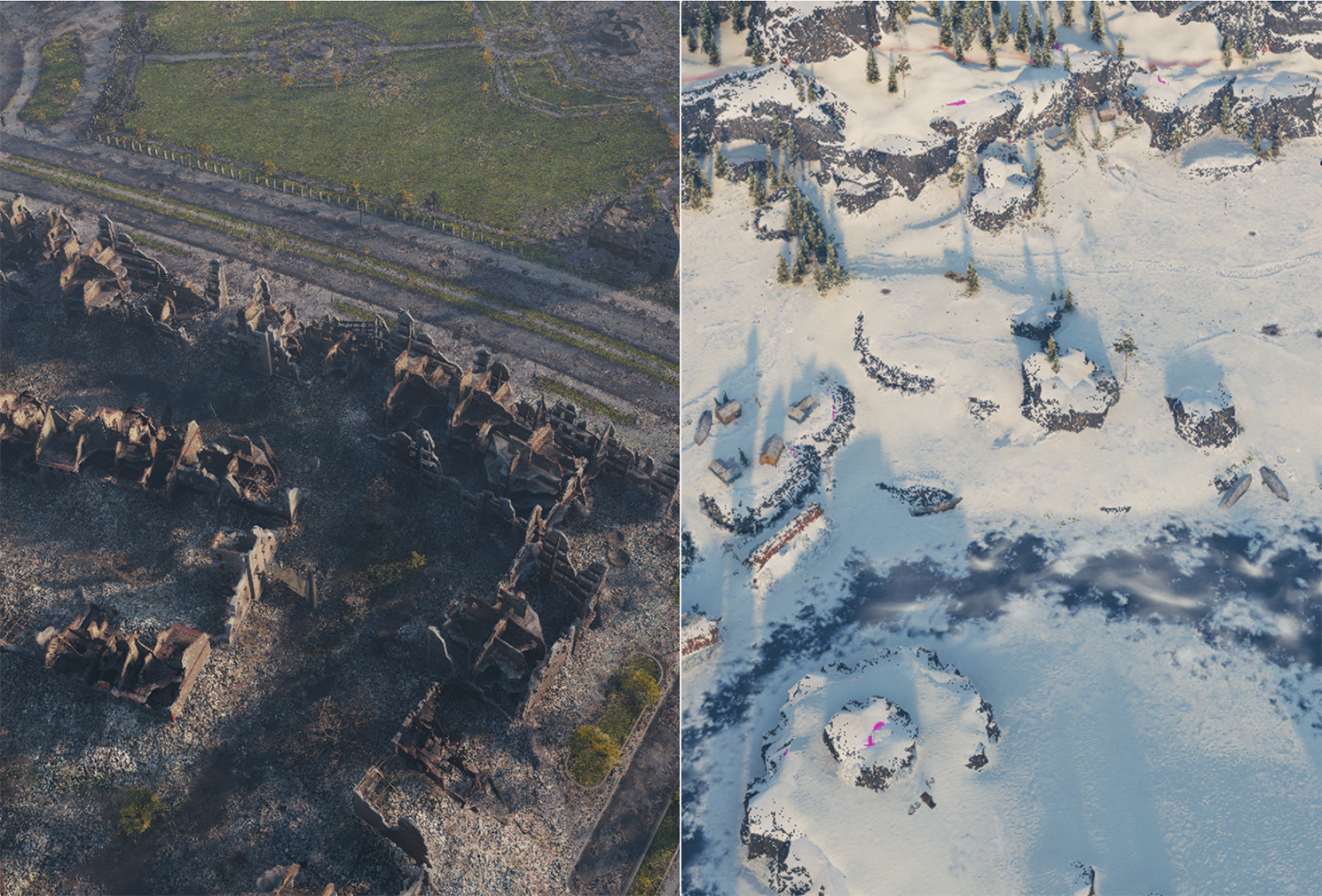}
    \caption{Examples of real map environment from \emph{World of Tanks}~\protect\cite{game:wot}.}
    \label{fig:fig2}
\end{figure}

Video games are a content-intensive medium, requiring different types of assets. A demand which even has increased over time due to technological advancements (e.g., high-fidelity graphics, large screen resolutions) itself but also due to game environments becoming larger and more diverse. As such, different techniques for procedural content generation (PCG), i.e., \emph{the automatic or computer-assisted generation of content}~\cite{Togelius:2016} have been proposed to date with the goal of reducing development time and costs. With the emergence of advanced generative artificial intelligence (AI) techniques, research on PCG has flourished further and facilitating the creation of richer and more complex content. By now manifold PCG solutions for different and diverse assets such as textures~\cite{Murphy:2020}, and 3D characters~\cite{Li:2023} and sprites~\cite{Coutinho:2022}, or even of complete tile maps~\cite{Merino:2023}, levels~\cite{Shaker:2011}, or quests~\cite{deLima:2022} and more exist. Such content can be generated offline or online~\cite{Togelius:2011}, that is in advance or dynamically while a player is playing the game (e.g.,~\cite{Hastings:2009}). In this work we make use of generative techniques for assisting in the design of game maps, a crucial but also complex and labor-intensive task in game development. Artists must carefully craft landscapes, structures, and environmental details to ensure both aesthetic quality and gameplay coherence. Traditionally, this process requires scrupulous manual work, where every element is drawn and refined by hand. While this approach provides maximum artistic control, it is time-consuming and inefficient, especially for large-scale game worlds or iterative design workflows. As game environments grow increasingly complex (see Figure~\ref{fig:fig2}) as mentioned above, new tools are needed to support artists in accelerating the creation process without sacrificing artistic quality. 

However, fully automated solutions often lack the nuanced control that artists require, leading to results that feel generic, inhomogeneous, or disconnected from the desired artistic vision. As such, our proposed technique does not aim for a fully automated solution but which rather uses generative AI as an assistive tool where the human remains in the loop. Such a humans-in-the-loop approach harnesses to computational efficiency of computers with the judgment, expertise, and emotional sense of humans (see, e.g.,~\cite{Kumar:2024}). Thus, to bridge the gap between automation and artistic intent, we propose a hybrid approach: a smart editing brush that assists artists by intelligently filling user-defined areas of the game map (see Figure~\ref{fig:teaser}), thus approaching the task from an inpainting perspective (i.e. the act of filling in missing areas of an image~\cite{Bu:2024}) but also considering from scratch generation. For this purpose, we propose two variations, one brush based on generative adversial networks (GANs)~\cite{Goodfellow:2014} and one on latent diffusion models~\cite{Rombach:2022}. Instead of manually designing every detail from scratch, artists can use our brush to generate a preliminary layout -- on which to potentially further built upon -- within seconds. Our algorithm fills in user-specified areas with contextually relevant content, which can then be further refined by the artist. In doing so, our AI-assisted \emph{Smart Brush} aims to enhance the efficiency and quality of 3D gaming map art creation by enabling seamless, high-resolution texture manipulation. Rather than replacing artists, it aims to empower them by reducing the time spent on repetitive groundwork while preserving creative freedom. Additionally, the proposed tool is supposed to hold significant business value through optimizing content creation and reducing production time. This, in turn, can contribute to cost savings and increased efficiency in game development pipelines. In summary, the key contributions of our work are:

\begin{itemize}[leftmargin=*]
    \item Artist-centric AI-based system: We can adapt seamlessly generated textures for a variety of map materials and different game genres, making the technique broadly applicable. Unlike traditional procedural methods that generate content in isolation, our approach preserves artistic intent and integrates AI-assisted automation, bridging the gap between full automation and manual artistic control.
    \item Both high-fidelity inpainting of missing regions and full content generation from scratch: Our dual approach (\emph{BrushGAN} and \emph{BrushCLDM}) manages both partial masks and cases where the entire region is covered with mask.
    \item Multi-chunk stitching mechanism for large-scale map generation: We incorporate a stitching mechanism that guarantees invisible boundaries between chunks (fundamental map units).
    \item Advanced context injection: By leveraging multi-stage training, material templates, and relevant context, our system produces highly organic textures – even in challenging pure generation scenarios (when the whole image is generated). The template-guided color coherence method ensures that generated textures maintain consistent color and structure, even in cases of completely missing regions.
    \item A novel hybrid pipeline for seamless texture generation: We are the first to provide a comprehensive pipeline, combining the most modern generative approaches trained with context-aware material integration, and seamless texture synthesis with hand-crafted algorithms for multi-chunk stitching, ensuring high-quality, artist-like results across diverse map materials and game environments.
    \item Through comprehensive evaluations, we demonstrate that our method outperforms state-of-the-art alternatives (\emph{SPADE}~\cite{Park:2019}, \emph{SN-PatchGAN}~\cite{Yu:2019}, \emph{Palette}~\cite{Saharia:2022}, \emph{CLDM}~\cite{Zhang:2023}, etc.) across multiple inpainting and full-generation scenarios. Our system achieves higher FID~\cite{Heusel:2017}, LPIPS~\cite{Zhang:2018}, and SSIM~\cite{Wang:2004} scores, showcasing superior realism, perceptual quality, and structural consistency.
\end{itemize}

\noindent In the remainder of this paper, we first review related works on AI-assisted art tools (Section~\ref{sec:relatedwork}), detail our methodology (Section~\ref{sec:onechunk} and Section~\ref{sec:multichunk}), and evaluate the tool's effectiveness through practical use cases (Section~\ref{sec:evaluation}) before concluding the paper in Section~\ref{sec:conclusions}.

\section{Related Work}
    \label{sec:relatedwork}

The full range of PCG is too broad to cover here in detail. As such we focus our review on works related to automated map manipulation tools and generative modeling with respect to game maps. In particular, since our system builds upon and extends prior research in inpainting~\cite{Liu:2018,Elharrouss:2020} and generative methods~\cite{Yang:2023b,Deshmukh:2024} we will put emphasis on this line of research. 

\subsection{Inpainting and Image Generation}

In recent years, deep learning methods have taken center stage in the field of image synthesis and generation, particularly GANs~\cite{Goodfellow:2014,Wang:2018,Radford:2016,Karras:2018,Brock:2019}, diffusion models~\cite{Dhariwal:2021,Shah:2024,Yang:2023a,Song:2022}, and more advanced convolutional neural networks~\cite{Morrow:2020,Vahdat:2020,Parmar:2018,Cai:2024}. A notable milestone in image inpainting was the introduction of Free-Form Gated Convolutions (\emph{SN-PatchGAN})~\cite{Yu:2019}, which allowed for flexible inpainting with improved feature learning for irregularly shaped missing regions. \emph{StyleGAN}~\cite{Karras:2019}, for instance, designed the generation process via progressive growing and style-based latent spaces, leading to high-resolution outputs. \emph{SPADE}~\cite{Park:2019} introduced spatially adaptive normalization layers, significantly enhancing spatial coherence and semantic fidelity in image synthesis. More recently, \emph{Palette}~\cite{Saharia:2022} utilized diffusion-based techniques to achieve diverse and high-fidelity image-to-image transformations, addressing limitations of prior GAN-based models in capturing output variability.

Likewise, high-resolution Latent Diffusion Models (LDM)~\cite{Rombach:2022} employ a diffusion process in a compressed latent space to efficiently generate high-fidelity, high-resolution content, making them especially well-suited for tasks requiring detailed texture synthesis and inpainting. 

Advancements in AI-supported image inpainting have not only led to solutions aimed towards general image inpainting (e.g.,~\cite{Yu:2023}) but also to their application in specific domains such as forensics~\cite{Zhu:2023}, manga drawings~\cite{Xie:2021}, or medicine~\cite{Tran:2021} to name but a few. Inpainting has also found first applications in relation to video games although its adaption is currently rare. Gonzalez and Guzdial~\cite{Gonzalez:2023} focused on level inpainting, that is for reconstructing and extending video game levels, by using a \emph{U-Net} as well as an \emph{Autoencoder} and using \emph{Super Mario Bros.} as a use case. 

Building on these advancements, we introduce a novel system offering a dual approach: \emph{BrushGAN} and \emph{BrushCLDM} that operate not only in the inpainting scenario but also in a pure generation scenario. Both approaches are guided by additional contextual data about the gaming map with updated multi-stage training scheme. By focusing on maps, our inpainting approach operates on multiple input images (masks), in contrast to usual image inpainting methods, which focus on a single image.

\subsection{PCG and Map Creation}

PCG for game map creation is a well-studied topic. Works such as by Togelius et al.~\cite{Togelius:2016} and techniques such as \emph{WaveFunctionCollapse}~\cite{Merrell:2009} showcase automated workflows for level design and texture synthesis. However, such 'traditional' PCG methods often lack the flexibility and high-resolution detail demonstrated by modern deep learning approaches. For example, Mao et al.~\cite{Mao:2024} provide a comprehensive survey on how generative artificial intelligence is applied to PCG across multiple domains, including gaming and simulation. The study highlights the advantages of deep learning models such as GANs and diffusion models over traditional PCG methods, particularly in terms of adaptability and quality.

At this point it is also worth nothing, that the term 'map generation' can be related to geometric terrain generation and to image-based techniques such as creating tile maps or painting the terrain's surface. Approaches to terrain generation often make us of different established noise functions such as Perlin~\cite{Hyttinen:2017} or Worley noise~\cite{Howard:2022}. However, these might look unnatural and unappealing and lack controllability~\cite{Nunes:2023}. As such generative approaches are also being increasingly used in this area as well. For instance, Lochner et al.~\cite{Lochner:2023}, albeit not concerned with games specifically, focused on generating heightfield terrains which serve as foundation for computer-generated natural scenes. Their framework incorporates diffusion models for conditional image synthesis, trained on real-world elevation data. Consequently, its applicability is limited to elevation-based landscapes such as cliffs, canyons, and plains. With maps for strategy games in mind, Nunes et al.~\cite{Nunes:2023} explored the usefulness of different variants of GAN-based networks for heightmap generation. Such heightmaps, basically being images, are then used to derive the geometry from them.

However, our work is not concerned with creating the geometry but rather operates on an image level. Work in this direction also bifurcates into different directions such as the automatic generation of tile maps (e.g.,~\cite{Gabriel:2024,Kaczmarzyk:2023}) or more recently, but less pronounced to date, the generation of contiguous maps that are not composed of individual tiles. With respect to the latter, Siracusa et al.~\cite{Siracusa:2021}, for instance, proposed a method for blending GAN-generated textures to synthesize high-resolution 2D town maps for role-playing games (RPGs). Their approach focuses on seamless texture blending across different map regions, similar to our multi-chunk texture generation approach, ensuring a cohesive and artistically consistent final result.

Tang and Sweetser~\cite{Tang:2023} developed a GAN-based approach to generate multilayer maps of terrain based on statistical attributes that are provided as input. Their approach considers a height layer and a segmentation layer that indicates ground categories (e.g., water, grass). In contrast, our approach takes multiple masks -- defining the blending of materials -- as input. 

Lastly it should be mentioned that while many approaches are purely automated and lack control, many also recognize the importance of human involvement. For instance, the above cited work by Kaczmarzyk and Frejlichowski~\cite{Kaczmarzyk:2023} generates maps based on a set of input tiles as well as a user-provided sketch outlining the intended structure. Similarly, Tang and Sweetser~\cite{Tang:2023} considered input conditions to provide human control and Matthews and Malloy~\cite{Matthews:2011} proposed a system that build maps with cities and towns scattered across in such a way that they meet developer constraints (e.g., distances). 

Our work is also different from these in a sense that we are not concerned about generating whole maps from scratch but refining the visual appearance of existing ones. Using the concept of a brush we aim to provide an apt analogy to the art of painting. In doing so, our proposed solution handles both texture synthesis and diverse map materials to ensure seamless integration across various game environments and artistic styles. In this sense, our work also shares some resemblances with AI-driven texture mapping solutions like the work of Yin and Song~\cite{Yin:2024}  for texturing 3D models. Similar to our work they utilized \emph{Stable Diffusion} and GAN-based architectures to improve high-resolution texture generation. While focused on a different form of assets with different requirements, their goals of enhancing texture coherence and reducing manual labor in asset creation closely align with our aim of generating high-quality game maps.

\begin{figure}[t]
    \centering
    \includegraphics[width=0.9\linewidth]{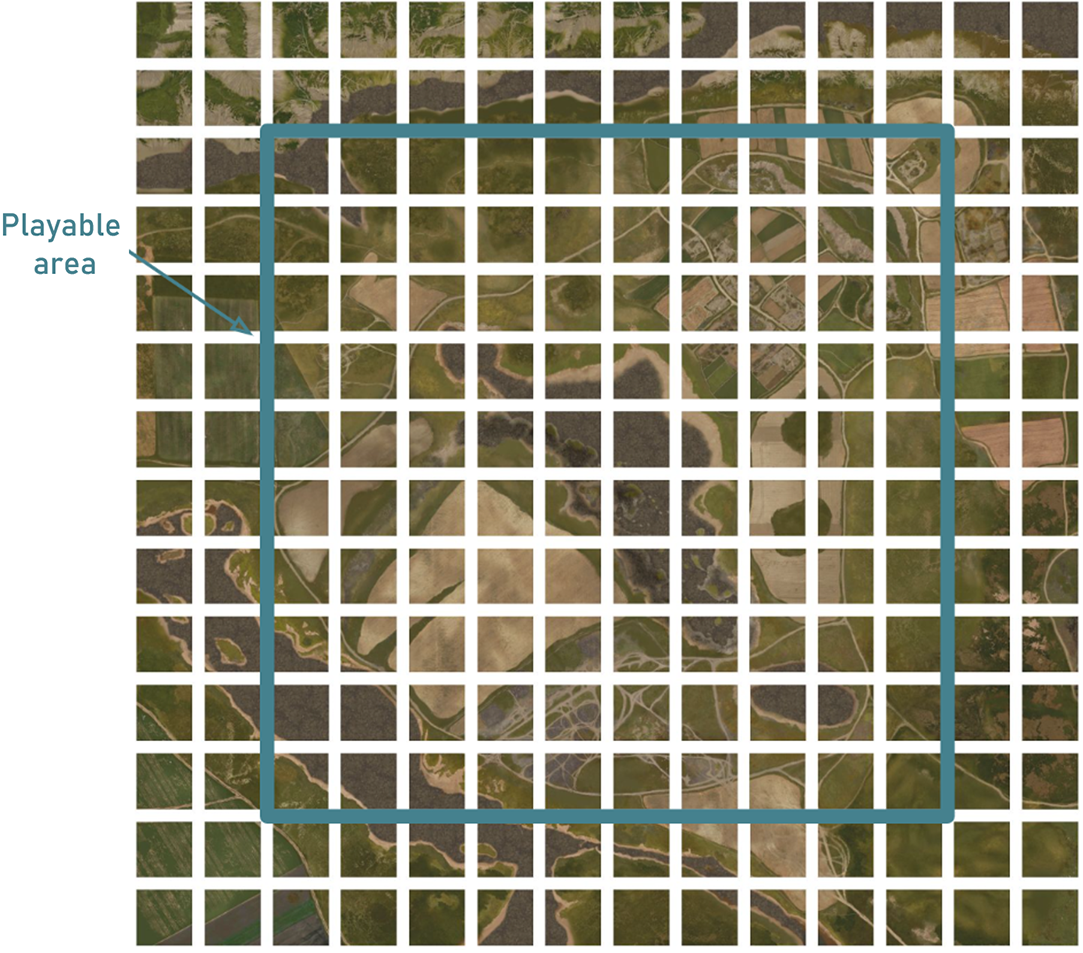}
    \caption{A \emph{WoT} map structurally consists of a 2D grid of chunks, usually divided into $10 \times 10$ chunks inside the playable area.}
    \label{fig:fig3}
\end{figure}

\subsection{Summary}

The above studies highlight the growing role of generative AI in game map generation and texture synthesis, further motivating our proposed \emph{Smart Brush} approach, which extends these concepts to gaming map creation with advanced context-aware generation models. In comparison to related approaches, our \emph{Smart Brush} combines modern generative models with handcrafted multi-chunk stitching algorithms to provide flexibility, realism, and adaptability across diverse game environments.

\section{Data Representation in Gaming Maps}   
    \label{sec:maps}

In the following we describe the structure of gaming maps on which our proposed methods operate on as well as the dataset used for learning and performance evaluation.

\subsection{Structure of Gaming Maps}

\begin{figure}[t]
    \centering
    \includegraphics[width=0.8\linewidth]{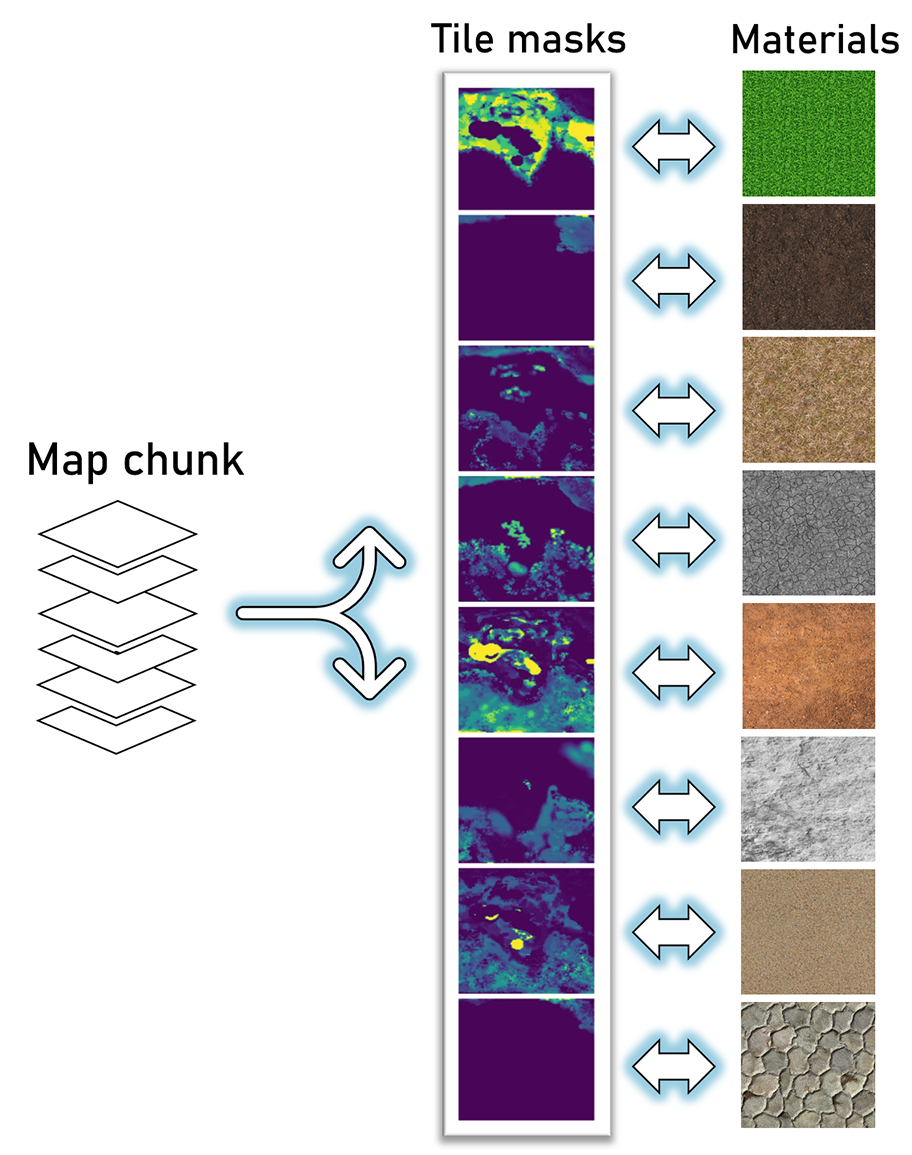}
    \caption{Each chunk of the map consists of 8 tile masks with their corresponding texture materials. The tile masks define the blending ratios of the materials.}
    \label{fig:fig4}
\end{figure}

In this work, we utilize maps from \emph{World of Tanks} (\emph{WoT})~\cite{game:wot} -- a team-based multiplayer online game -- as a reference for our map generation approach, where large-scale, high-detail environments and visual realism play a crucial role for the gameplay. In \emph{WoT}, a typical gaming map consists of a grid of chunks arranged in a regular $10 \times 10$ grid layout (see Figure~\ref{fig:fig3}) that comprise the playable area. Every chunk is further divided into 8 tile masks, each of which is associated with a specific texture material from a predefined database. These materials include common surface types such as soil, grass, ground, stone, and others (see Figure~\ref{fig:fig4}). Each material, in turn, is represented as a 3-channel RGB texture, while each tile mask is a 1-channel grayscale image that determines the blending ratio of the corresponding material in the final rendered map.

In other words, the main purpose of the tile masks is to smoothly blend multiple materials to achieve a realistic map appearance. The final appearance of the map is computed as a linear combination of the materials weighted by their corresponding tile masks, that is:

\begin{equation*}
    M_{final}\left(x,y\right) = \sum_{i=1}^{N} T_i \left(x,y\right) \cdot M_i\left(x,y\right)
\end{equation*}

\noindent where

\begin{itemize}
    \item $M_{final}\left(x,y\right)$ is the color at pixel $\left(x,y\right)$ in the final map, 
    \item $N$ is the number of materials (8 per chunk in our application scenario),
    \item $T_i\left(x,y\right)$ is the tile mask value (blending weight) for material $i$ at pixel $\left(x,y\right)$, and
    \item $M_i\left(x,y\right)$ is the texture value of material $i$ at pixel $\left(x,y\right)$.
\end{itemize}

\noindent Each pixel in the final map is obtained through this blending function, ensuring that transitions between different materials appear smooth and natural. Figure ~\ref{fig:fig5} (left) provides an example of the appearance of the final map together with a coarse visualization of the dominant materials.

It is important to note, that the amount of data in a gaming map is enormous. With a $10 \times 10$ chunk layout, each containing 8 tile masks, the total number of individual tile masks per map is 800. Since each tile is manually drawn by an artist, this process is highly time-consuming and labor-intensive. Automating this process with generative AI-based approaches such as our proposed \emph{Smart Brush} can significantly reduce manual workload while maintaining high-quality results.

\subsection{Dataset}
    \label{sec:dataset}

\begin{figure}
    \centering
    \includegraphics[width=1.0\linewidth]{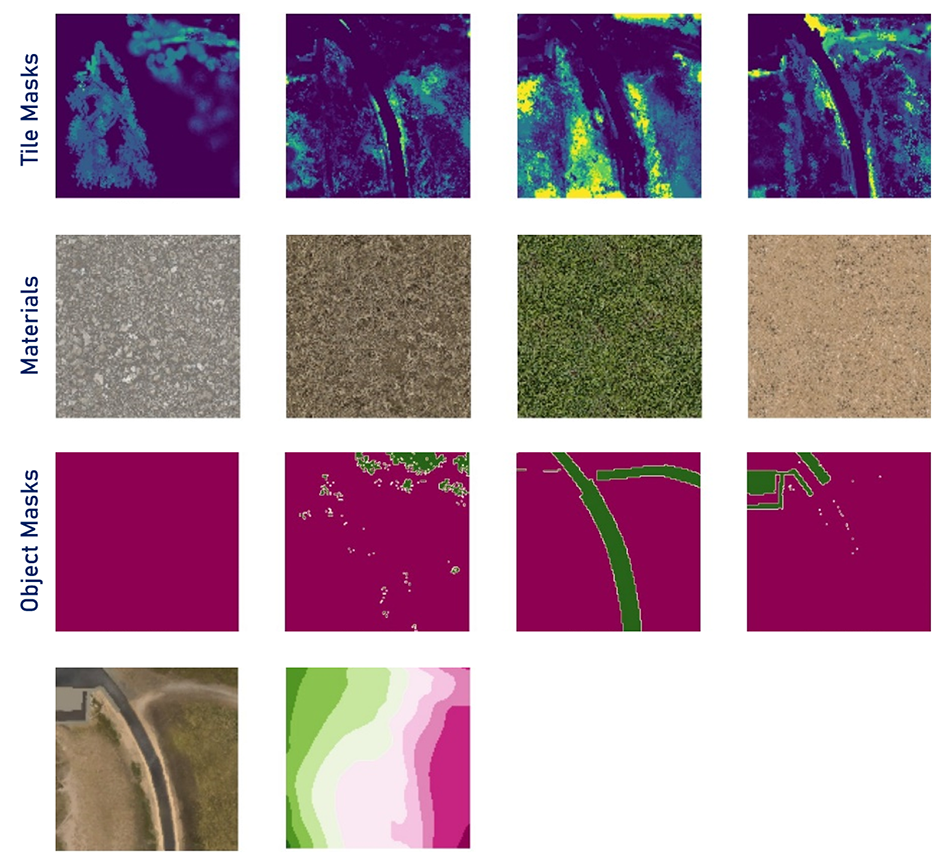}
    \caption{Examples of extracted map components: The top row shows a subset of tile masks, the second row their corresponding materials, and the third row provides examples of object masks (water, trees, roads, and buildings). The bottom row displays the global AM (left image) as well as a height map (right image).}
    \label{fig:fig6a}
\end{figure}

Our dataset consists of 43 \emph{WoT} maps within three predefined categories: urban,
winter, and natural. For each chunk of each map in this dataset, we extracted the following components (also see Figure~\ref{fig:fig6a} which illustrates these components):

\begin{description}[leftmargin=*,itemsep=0.5\baselineskip]
    \item[Tile Masks:] Each chunk contains 8 tile masks that define the blending of the materials.
    \item[Material Set:] Each tile mask is associated with a specific material.
    \item[Global Albedo Map (Global AM):] An RGB image describing the fundamental map features such as fields, rivers, and roads, similar to satellite imagery.
    \item[Height Map:] A 1-channel image that represents the basic terrain relief of the map.
    \item[Labeled Object Masks:] Masks that identify the exact locations of objects such as buildings, trees, roads, and water.
\end{description}

\begin{figure}
    \centering
    \includegraphics[width=1.0\linewidth]{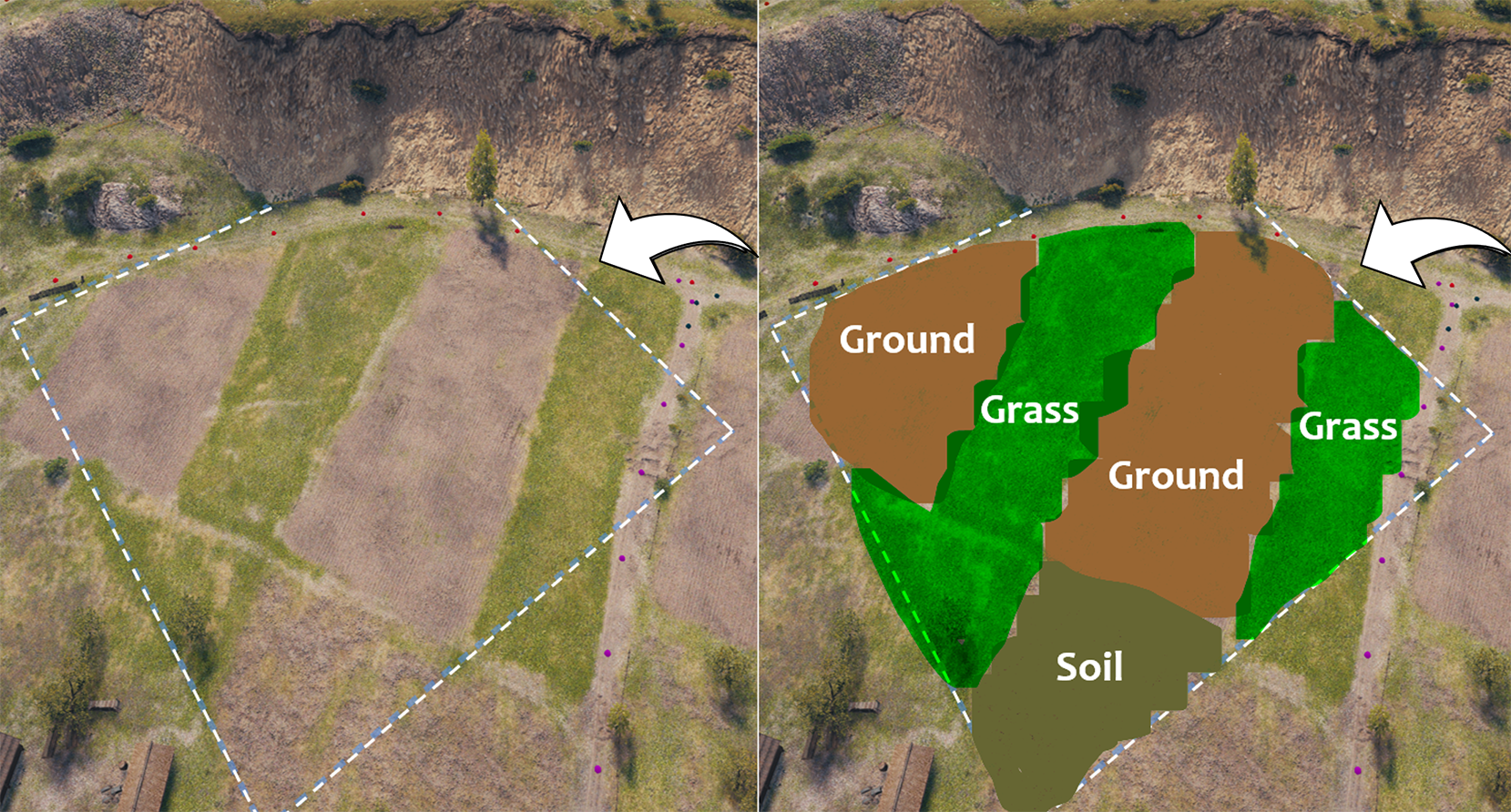}
    \caption{Left: Final map visualization obtained by blending the materials based on their tile masks for a chunk (white stippled line). Right: Rough visualization of the dominant materials.}
    \label{fig:fig5}
\end{figure}

\noindent Additionally, we employ on-the-fly stochastic generation of various mask types for inpainting scenarios. We will refer to these mask types as 'medium' masks that cover less than 30\% of the area, 'hard' masks that cover 30-99\%, and 'complete' masks that represent completely missing regions. These masks help in training models to handle a range of inpainting difficulties, from small missing patches to fully blank areas requiring complete reconstruction.

A crucial aspect of dataset preparation is ensuring that the data is properly split into training and test sets with similar distributions. To achieve this, we need a method to measure the similarity between two dataset distributions. For this purpose, we utilize the Fréchet Inception Distance (FID) score~\cite{Heusel:2017}. This metric evaluates the difference between two distributions by measuring the distance between feature representations of the datasets.

To ensure a balanced train-test split, we computed three key metrics:

\begin{itemize}[leftmargin=*]
    \item Pairwise map FID similarity ranking based on Global AM.
    \item Pairwise map FID similarity ranking based on tile masks.
    \item Material sets intersection percentage between maps.
\end{itemize}

\noindent These metrics were selected because the Global AM represents color-based differences in the maps, tile masks define the artistic style on the generated content, and material sets intersection helps to have a similar material structure between the compared maps. The final score $S$ for splitting is the average of these three metrics, mathematically: 

\begin{equation*}
    S = \frac{FID_{global\_am} + FID_{tiles} + P_{material}}{3}
\end{equation*}

\noindent where $FID_{global\_am}$ is the FID distance for Global AM instances, $FID_{tiles}$ is the FID for tile masks, and $P_{material}$ is the material sets intersection percentage.

\begin{figure}
    \centering
    \includegraphics[width=1.0\linewidth]{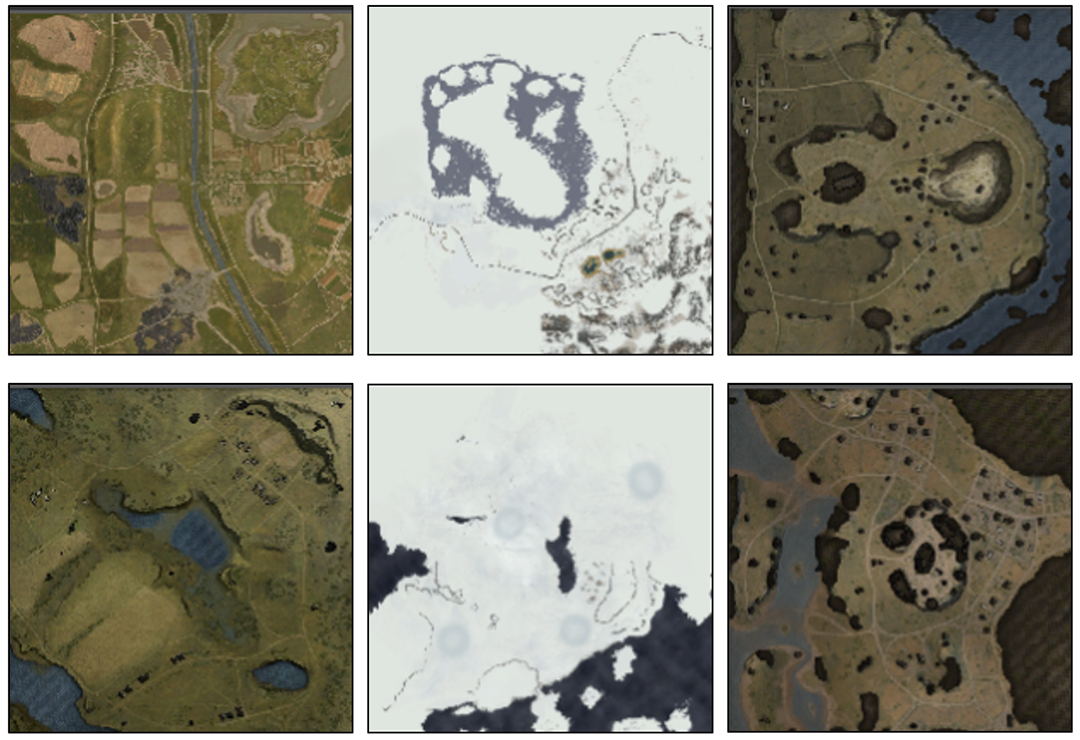}
    \caption{The most similar maps based on computed score. The upper row is moved to the training set, the bottom row – to the test set.}
    \label{fig:fig6b}
\end{figure}

For our dataset, we aim to find the three most similar map pairs within each of the three predefined categories based on the computed scores (see Figure~\ref{fig:fig6b}). As a result, three maps are fully moved to the test set to ensure test distribution aligns well with the train set. The validation set is prepared using the same approach to maintain consistency. This method ensures that our test data is highly representative of the training distribution, allowing for more reliable model evaluation.

\section{One-Chunk Brush Generation}
    \label{sec:onechunk}

For one-chunk generation with medium and hard modes, the goal is to generate the region inside the eight tile masks that are covered by the brush mask while maintaining seamless transitions between textures. This requires the model to reconstruct missing details within each affected tile while ensuring consistency with neighboring tiles and preserving the overall structural and stylistic integrity of the map. We address the task by proposing two separate methods (\emph{BrushGAN} and \emph{BrushCLDM}) build upon GANs and diffusion models. GANs are widely recognized for their faster training and inference as well as their ability to produce fine-grained, high-resolution textures, whereas diffusion models provide higher variability and better global consistency (cf.~\cite{Dhariwal:2021,Kuznedelev:2024,Kang:2025}). 

\subsection{BrushGAN}  
    \label{sec:brushgan}

\begin{figure}
    \centering
    \includegraphics[width=1.0\linewidth]{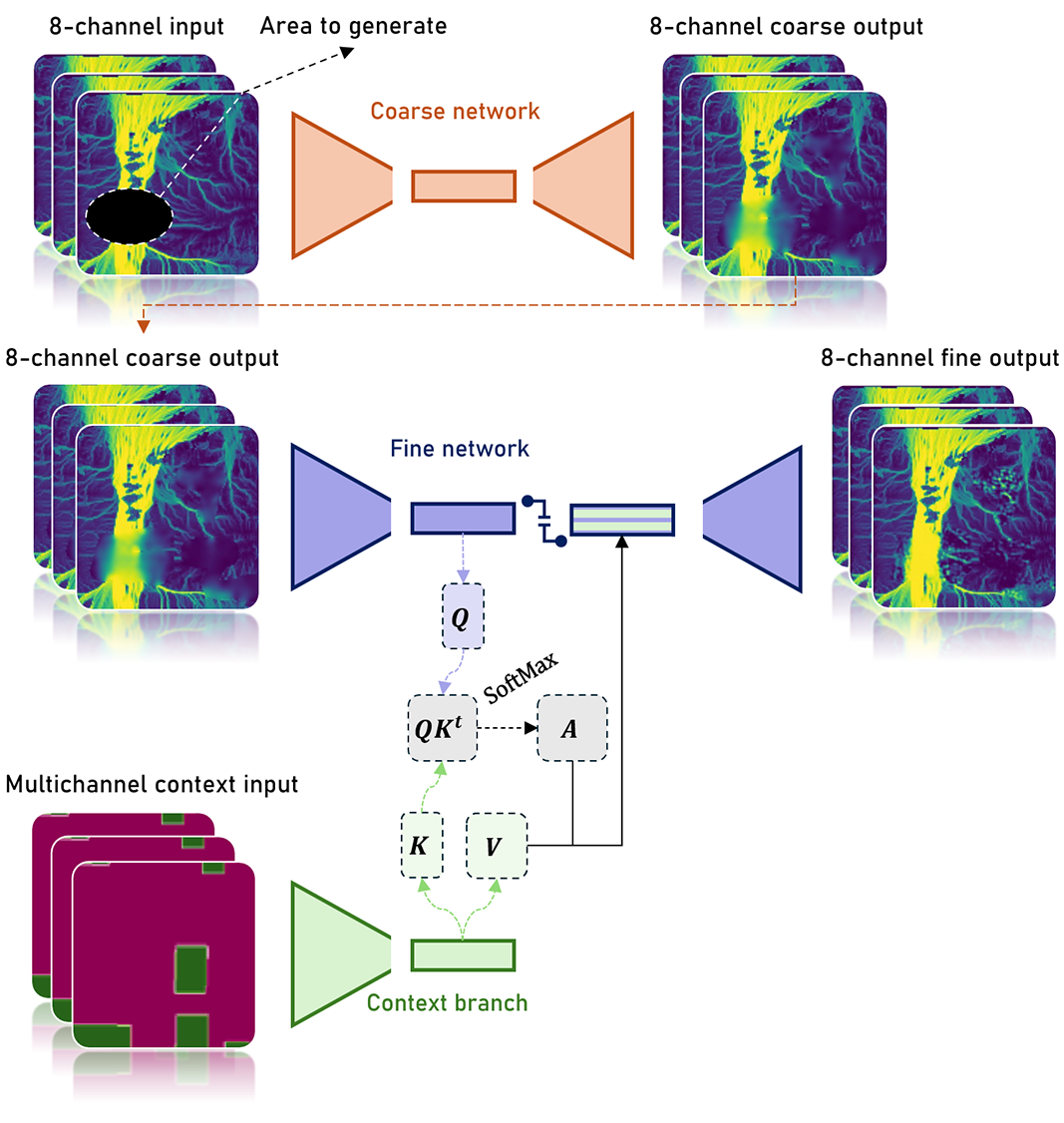}
    \caption{A visualization of our \emph{BrushGAN} architecture which consists of two autoencoders: a coarse autoencoder responsible for generating the initial structure and a fine autoencoder that refines details. To enhance contextual understanding, the fine autoencoder integrates additional context information (e.g., Global AM, height maps, and masks) using a cross-attention mechanism to ensure improved coherence and texture consistency.}
    \label{fig:fig7}
\end{figure}

The \emph{BrushGAN} approach is based on \emph{SN-PatchGAN}~\cite{Yu:2019} and employs a two-stage structure~\cite{Cai:2019}. The first stage consists of a coarse autoencoder that generates an initial approximation of the missing region, while the second stage is a fine autoencoder that refines texture and structural details. A gated convolution mechanism is incorporated to learn a dynamic feature selection mechanism for each channel at each spatial location across all layers, improving the color consistency and inpainting quality.    

A key distinction from prior work by Yu et al.~\cite{Yu:2019} is in the training data. While the original method operates on standard RGB images, our approach processes an 8-channel input, where each channel corresponds to a different material layer. \emph{BrushGAN} takes 8 tiles as input alongside the brush mask and restores a multi-channel (8-channel) image, ensuring simultaneous restoration of multiple texture layers and maintaining material consistency across different map elements.

\begin{figure*}[t]
    \centering
    \includegraphics[width=1.0\linewidth]{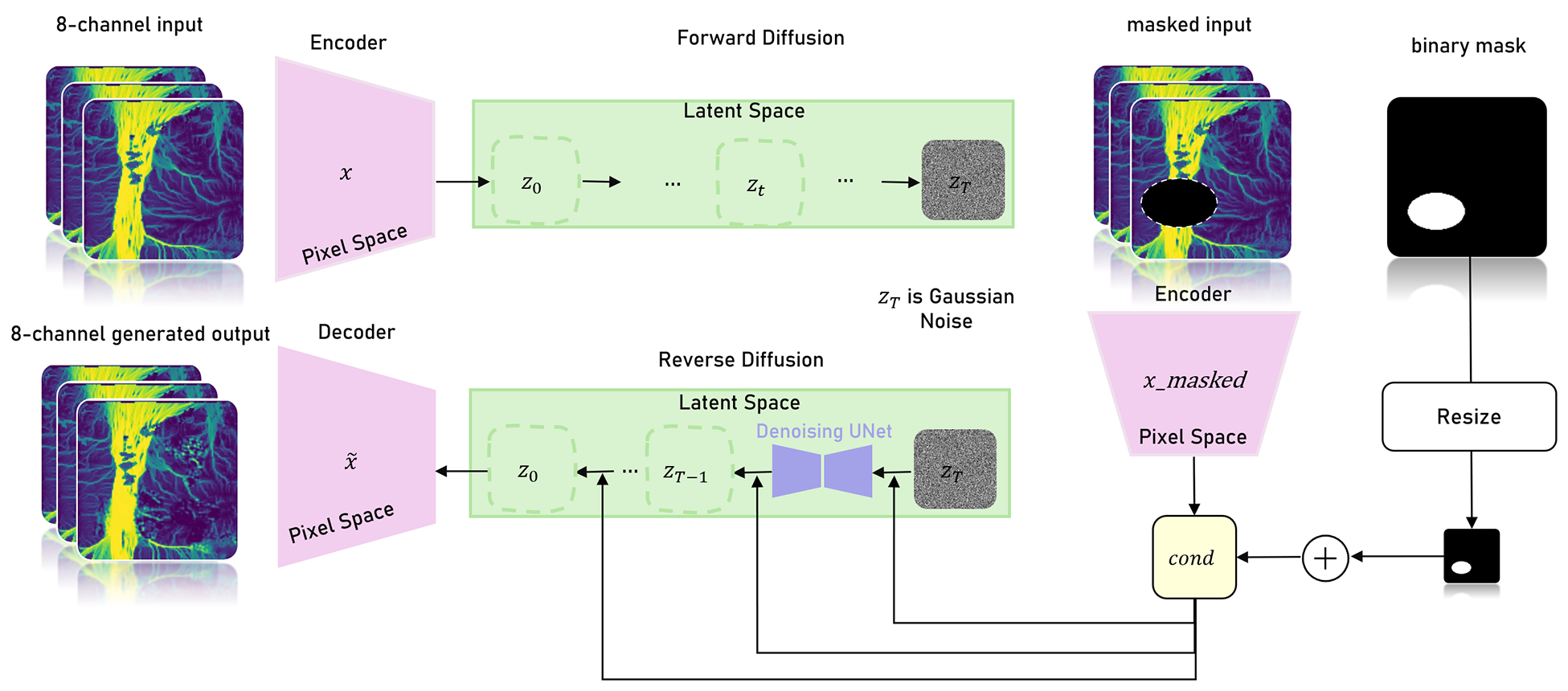}
    \caption{A visualization of our \emph{BrushCLDM} generator architecture. It processes eight tile masks through an encoder, generating corresponding latent representations. These undergo a forward diffusion process, converting them into Gaussian noise. The model is conditioned on a latent-masked input, encoded similarly and resized to match the latent space. The reverse diffusion process then removes noise over \emph{T} steps, producing a denoised latent representation, which is finally decoded back into pixel space to generate the output image.}
    \label{fig:fig8}
\end{figure*}

The generator is trained using a multi-component loss function. We employ mean squared error (MSE) loss for both coarse and fine stages and \emph{VGG-19}~\cite{Simonyan:2015} perceptual loss~\cite{Johnson:2016} for the fine stage. Additionally, adversarial loss is used in the fine stage, as described by Yu et al.~\cite{Yu:2019}. 

However, unlike Yu et al.~\cite{Yu:2019}, our training process follows a progressive learning strategy. Initially, we train the coarse generator, allowing it to learn a stable structural representation of the missing regions. After a certain number of epochs, we freeze the coarse generator and shift the focus entirely to the fine generator which refines texture details and enhances realism. This step-by-step approach helps the network to progressively learn from coarse structures to fine details, rather than optimizing both simultaneously. The discriminator is utilizing the \emph{PatchGAN}~\cite{Li:2016} architecture to assess the quality of the generated textures.

In addition, in contrast to the original model, we integrate additional spatial context in the form of Global AM, height maps, and object masks (see Figure~\ref{fig:fig7}). To effectively incorporate these, we introduce a dedicated feature extraction branch that processes these inputs and integrates them into the bottleneck of the fine autoencoder using a cross-attention mechanism~\cite{Vaswani:2017}, defined as

\begin{equation*}
    F_{context} = softmax\left(\frac{QK^T}{\sqrt{d_k}}\right)V   
\end{equation*}

\noindent where 

\begin{itemize}
    \item $Q$ (Query) represents the features of the main network, which need additional context to generate high-quality inpainting results. In our case, this comes from the bottleneck of the fine autoencoder in the \emph{BrushGAN},
    \item $K$ (Key) represents the contextual features from additional contextual branch, and
    \item $V$ (Value) contains the actual contextual information that will be fused into the main network.
\end{itemize}

\noindent This mechanism allows the model to extract global structure information, enhances texture synthesis, and ensures that inpainted regions align realistically with their surroundings. It reduces artifacts by leveraging additional spatial information and improves coherence between the restored and existing regions. By integrating multi-channel restoration, a progressive freezing training strategy, and context-aware cross-attention, our approach extends previous GAN-based inpainting methods to better handle gaming environments. These modifications allow for more accurate material blending, improved spatial coherence, and enhanced fine-detail synthesis.

\subsection{BrushCLDM}
    \label{sec:brushldm}

As a second generative method, we present \emph{BrushCLDM} which is built upon the concept of Latent Diffusion Models~\cite{Rombach:2022} and thus differs fundamentally from GAN-based approaches by relying on a diffusion process rather than adversarial training. While GANs often struggle with mode collapse~\cite{Srivastava:2017,Arjovsky:2017,Che:2017} and require careful balancing between the generator and discriminator, LDMs iteratively refine images through a controlled denoising process, leading to more stable training and higher-quality results~\cite{Rombach:2022}. Unlike traditional diffusion models such as DDPMs~\cite{Ho:2020}, LDMs process images within a compact latent space rather than working directly with high-dimensional pixel data.  

Following the original approach~\cite{Rombach:2022}, we leverage a noising-denoising diffusion process in the latent space and utilize a variational autoencoder (VAE)~\cite{Kingma:2022} for efficient encoding. However, we introduce several enhancements: (1) our VAE resolution is $64 \times 64$ which helps to reconstruct images with high quality, (2) the architecture has been reworked, resulting in an updated version with more than a $\times 6$ reduction in parameters compared to the original model, (3) we not only incorporate the losses proposed by Rombach et al.~\cite{Rombach:2022} but also additional losses, including perceptual loss~\cite{Johnson:2016}, focal frequency loss~\cite{Jiang:2021}, and style loss~\cite{Gatys:2016}.

An overview of the \emph{BrushCLDM} generator architecture is presented in Figure~\ref{fig:fig8}, while the following paragraphs provide a detailed explanation of each enhancement.

First, as noted above, VAE resolution has been changed to $64 \times 64$. The original LDM for inpainting, as provided by Rombach et al.~\cite{Rombach:2022}, operates with a $512 \times 512$ image resolution and a $32 \times 32$ latent representation. However, given our dataset and a tile size of $128 \times 128$, experimental results indicated that reducing the latent image resolution below $64 \times 64$ severely degrades the final output quality of the model.

Secondly, network optimization of the original architecture was conducted with \emph{WeightWatcher}~\cite{Martin:2024}, an open-source tool that analyzes weight matrices of deep neural networks to predict generalization performance. Results obtained through it enabled us to evaluate model scalability and enhance training stability. From the original model and its 387 M parameters we derived a smaller model with 60.5 M parameters.

With respect to the newly added losses, focal frequency loss~\cite{Jiang:2021} follows the idea that any image can be represented as a combination of frequencies with a Discrete Fourier transform, expressed mathematically as: 

\begin{equation*}
    L_{ffl} = \frac{1}{MN}\sum_{u=0}^{M-1}\sum_{v=0}^{N-1} w\left( u,v \right) | F_{gt}\left( u,v \right) - F_{gen}\left( u,v \right) |^2
\end{equation*}

\noindent where

\begin{itemize}
    \item $M, N$ are the dimensions of a 2D image,
    \item $\left( u,v \right)$ represent the coordinates of a spatial frequency on the frequency spectrum,
    \item $F_{gt}\left( u,v \right)$ is a Fourier representation of the ground-truth image,
    \item $F_{gen}\left( u,v \right)$ is a Fourier representation of the generated image, and 
    \item $w$ is a spectrum weight matrix which goal is to down-weight the easy frequencies and focus on hard ones~\cite{Jiang:2021}. 
\end{itemize}

\begin{figure}
    \centering
    \includegraphics[width=0.8\linewidth]{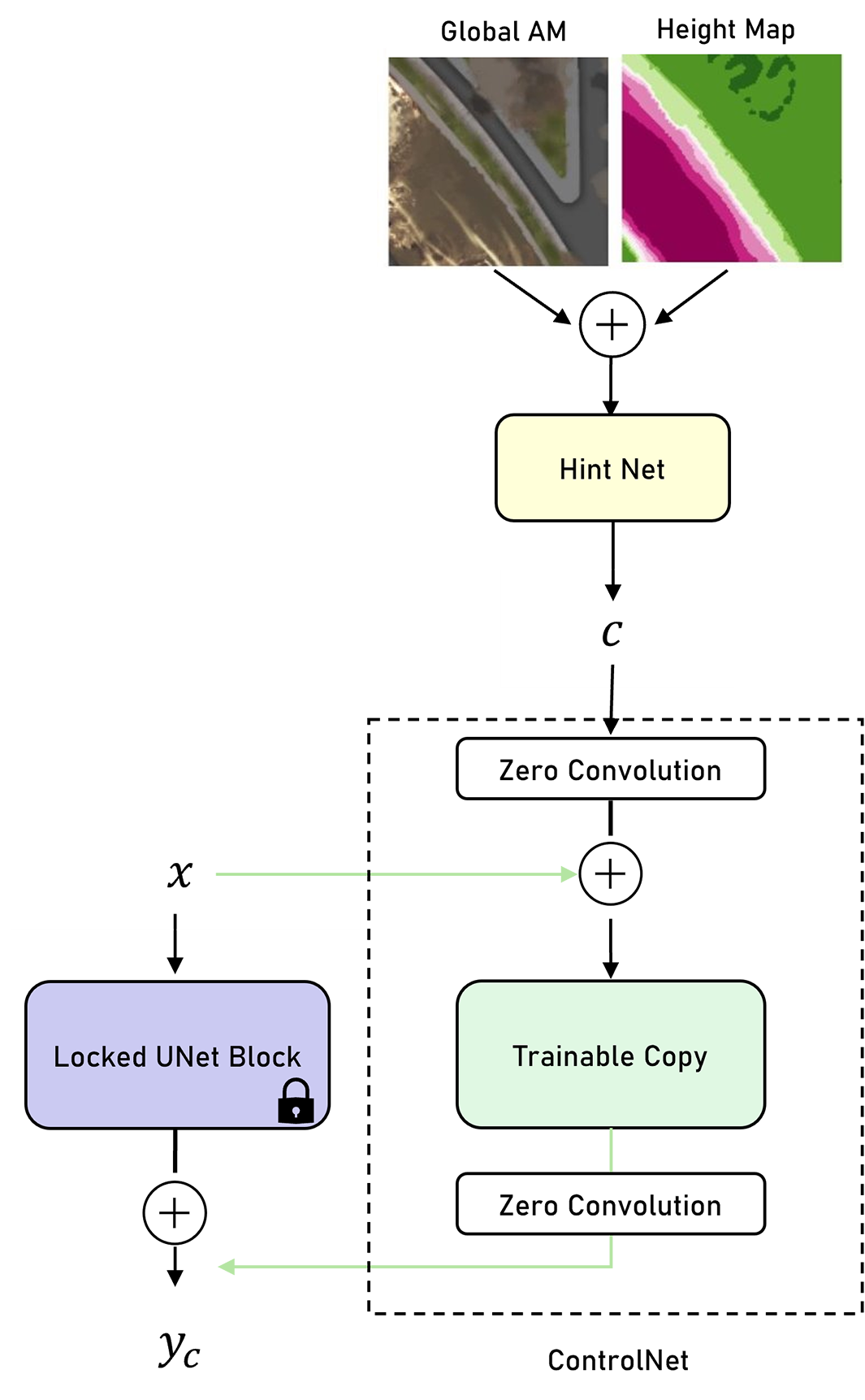}
    \caption{A visualization of the overall \emph{BrushCLDM} architecture. The hint network first processes the conditioning information, which is then integrated into trainable copies of the denoising \emph{U-Net} blocks via zero convolution. The original \emph{U-Net} blocks in generator part remain frozen, while \emph{ControlNet} is applied to introduce additional control over the generative process.}
    \label{fig:fig9}
\end{figure}

\noindent Focal frequency loss helps our model to preserve high-frequency details, thus yielding sharper generated images. Style loss~\cite{Gatys:2016}, on the other hand, is based on the calculation of Gram matrices, which capture the correlations between feature maps obtained with a \emph{VGG-19} convolutional neural network~\cite{Simonyan:2015}, ensuring that the generated image preserves the texture and artistic patterns of the style reference. The loss is defined as the squared Frobenius norm difference of the mentioned Gram matrices:

\begin{figure*}
    \centering
    \begin{minipage}[t]{0.27\linewidth}
        \centering
        \includegraphics[width=1.0\linewidth]{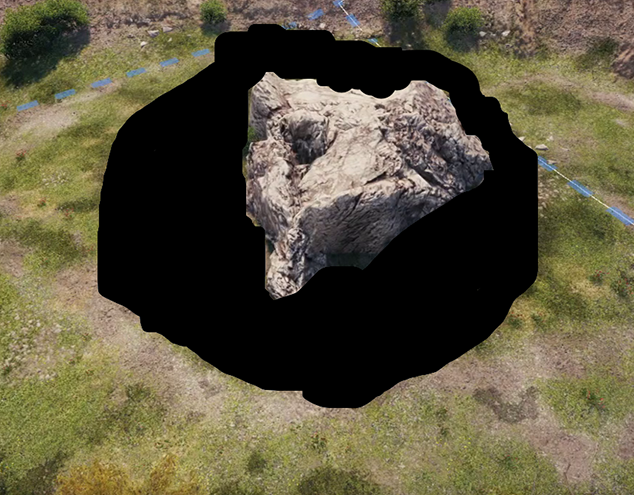}\\
        (a)
    \end{minipage}
    \hspace{0.5cm}
    \begin{minipage}[t]{0.27\linewidth}
        \centering
        \includegraphics[width=1.0\linewidth]{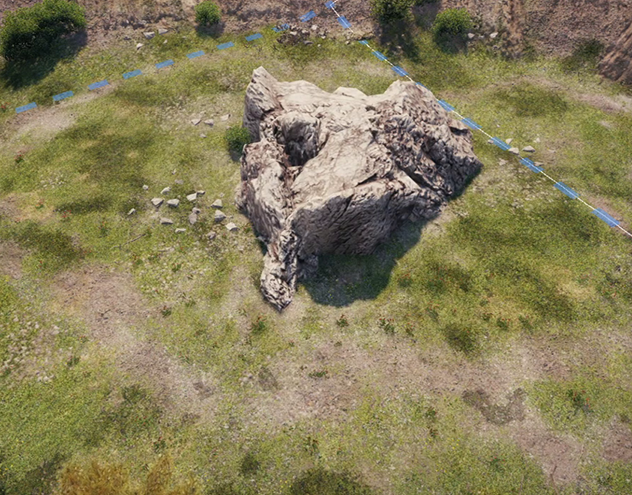}\\
        (b)
    \end{minipage}
    \hspace{0.5cm}
    \begin{minipage}[t]{0.27\linewidth}
        \centering
        \includegraphics[width=1.0\linewidth]{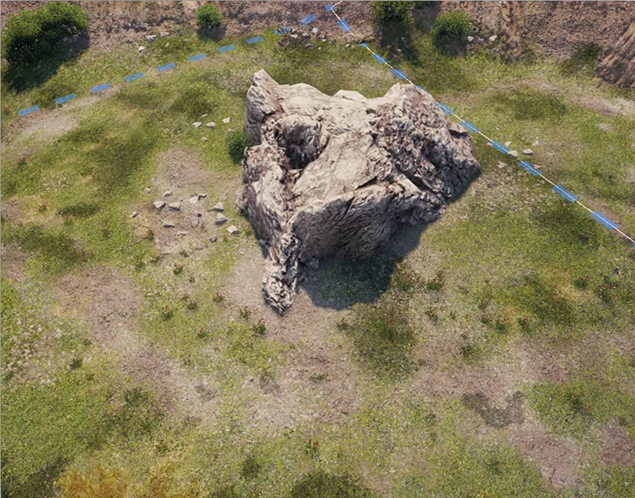}\\
        (c)
    \end{minipage}
    \caption{Example of \emph{BrushGAN} and \emph{BrushCLDM} functionality: (a) input with a masked region (black area), (b) output from \emph{BrushGAN}, and (c) result from \emph{BrushCLDM}.}
    \label{fig:fig10}
\end{figure*}

\begin{equation*}
    L_{style} = \sum_l w_l \frac{1}{4C_l^2H_l^2W_l^2} \sum_{i,j} \left( G_{gen\,i,j}^l - G_{style\,i,j}^l\right)^2
\end{equation*}

\noindent where

\begin{itemize}
    \item $w_l$ is a weight assigned to each layer's l contribution to the total style loss,
    \item $C_l$, $H_l$, $W_l$ are the number of channels, height and width of the feature map in the $l$-th layer,
    \item $i,j$ are the IDs of feature maps between which the correlation Gram matrix is calculated,
    \item $G_{gen\,i,j}$ is a Gram matrix calculated for a generated image between the $i$-th and $j$-th feature maps, and
    \item $G_{style\,i,j}$ is a Gram matrix calculated for the style image, in our case -- ground-truth image -- between the $i$-th and $j$-th feature maps.
\end{itemize}

\noindent In contrast to the \emph{Neural Style Transfer}~\cite{Gatys:2016} task, the style loss in our approach does not follow the goal of copying the style from one image to another but rather enforcing it to save global similarity in structural and style patterns. Putting everything together, our final loss $L_{total}$ is expressed as

\begin{equation*}
    L_{total} = \lambda_1 L_{mse} + \lambda_2 L_{perc} + \lambda_3 L_{ffl} + \lambda_4 L_{style}
\end{equation*}

\begin{figure}
    \centering
    \includegraphics[width=1.0\linewidth]{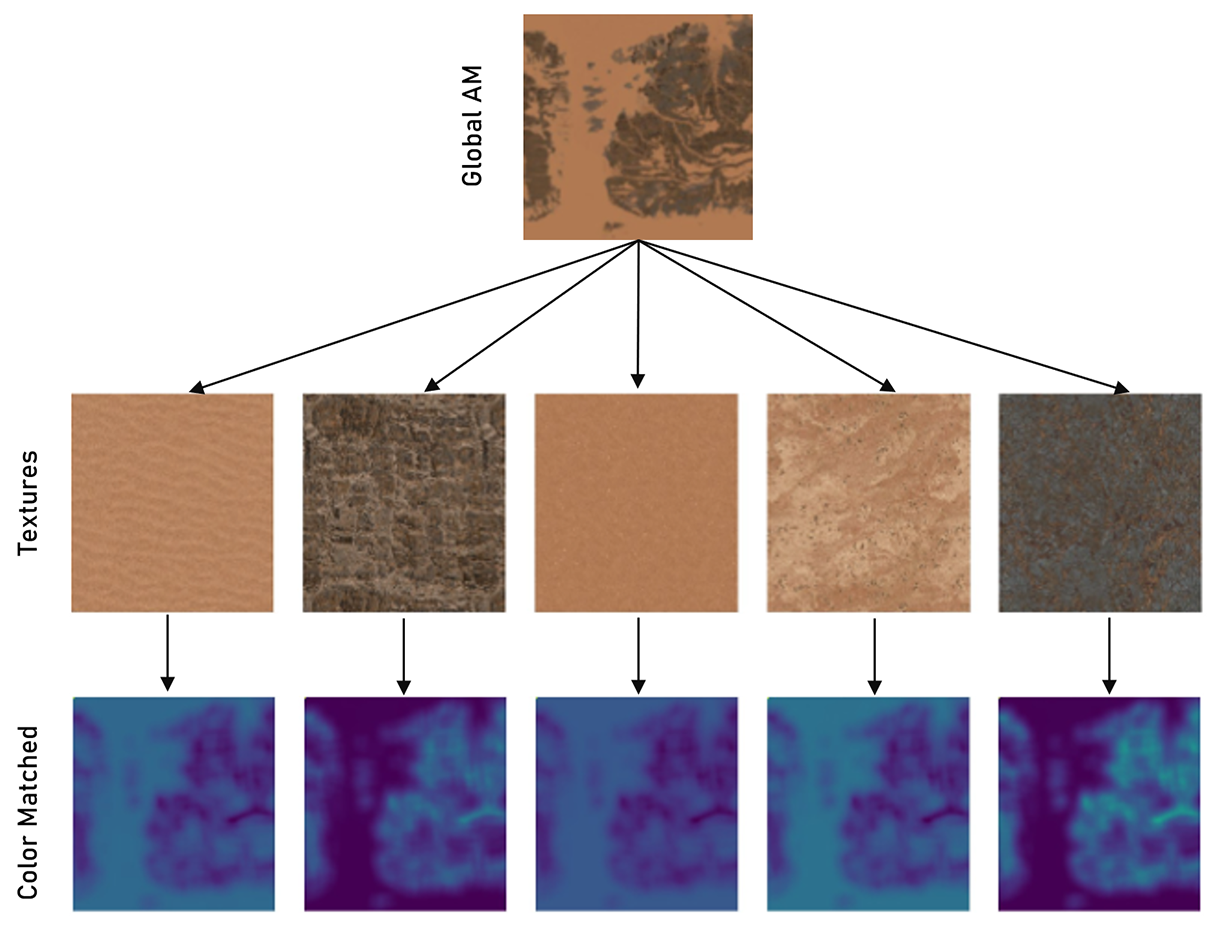}
    \caption{Top image: the Global AM. The second column shows the textures that make up the chunk, and the third column presents color-matched templates obtained with the Template-Guided Color Coherence algorithm.}
    \label{fig:fig11}
\end{figure}

\noindent where $L_{mse}$ is the Mean Squared Error between the noise predicted by our model and the actual added Gaussian noise and $L_{perc}$ is the perceptual loss~\cite{Johnson:2016}. $\lambda_1$ to $\lambda_4$ are the weights regulating the contribution of each individual loss term to the overall loss function.

Like in case of \emph{BrushGAN}, our next step was to leverage additional contextual information, further improving the quality of the generated outputs.

To do so we incorporate \emph{ControlNet}~\cite{Zhang:2023}, which introduces additional conditioning on external guidance, such as height maps and Global AM (see Figure~\ref{fig:fig9}). We are following the original implementation of the model with a few modifications, including the complete removal of text prompting, as our model does not use text as a conditioning input. Another small adjustment involved tweaking the strides of the tiny network responsible for extracting embeddings from the conditioning input to match the latent image. This structured conditioning approach enhances the model's ability to align the generated content with the underlying scene geometry, resulting in more coherent and context-aware outputs, particularly in complex scenarios.

Figure~\ref{fig:fig10} provides illustrative examples of results obtained via the presented \emph{BrushGAN} and \emph{BrushCLDM} architectures

\subsection{Whole-Chunk Generation}

\begin{figure*}
    \centering
    \begin{minipage}[t]{0.27\linewidth}
        \centering
        \includegraphics[width=1.0\linewidth]{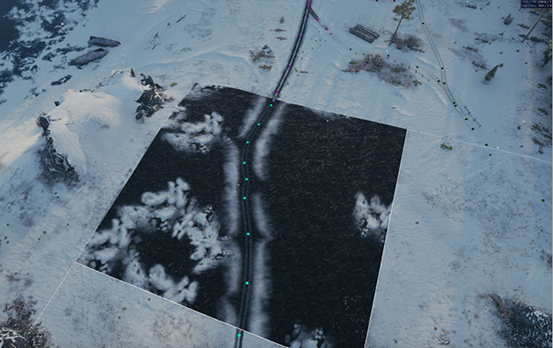}\\
        (a)
    \end{minipage}
    \hspace{0.5cm}
    \begin{minipage}[t]{0.27\linewidth}
        \centering
        \includegraphics[width=1.0\linewidth]{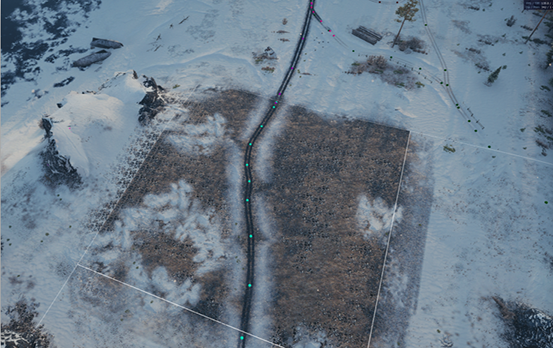}\\
        (b)
    \end{minipage}
    \hspace{0.5cm}
    \begin{minipage}[t]{0.27\linewidth}
        \centering
        \includegraphics[width=1.0\linewidth]{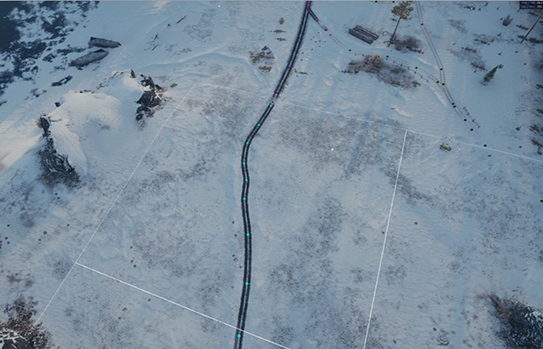}\\
        (c)
    \end{minipage}
    \caption{Effect of Template Matching on texture generation: (a) input with missing textures (black area), (b) generated result without template matching, and (c) result with template matching. The latter achieves better global color consistency, ensuring a more seamless integration with the existing map. }
    \label{fig:fig12}
\end{figure*}

Inpainting typically relies on the surrounding visual information to reconstruct missing regions in order to ensure consistency in textures, structures, and colors. However, when the entire area is masked, the model loses access to any contextual cues, making the generation significantly more challenging. This issue is further amplified in our case, as we work with 8 one-channel tiles, preventing the network from utilizing color information to maintain consistency across the generated outputs.

To address the challenges associated with whole-chunk generation, we propose two techniques (progressive training and template-guided color coherence) that are applied to both of our approaches:

\begin{description}[leftmargin=*,itemsep=0.5\baselineskip]
    \item[Progressive Training] employs a multi-stage training scheme, allowing the network to start with simpler tasks in the early epochs and gradually increase complexity over time. This approach enables the algorithm to adapt beyond relying solely on neighboring pixels for restoration, incorporating broader contextual information, even in the absence of neighboring data. 
    \item[Template-Guided Color Coherence] is a method which helps to identify dominant materials within a chunk. In the complete mask generation mode, where no neighboring pixels are available, the neural network lacks information to determine which tiles are important and which ones are not. To address this, we apply a template-matching technique to identify dominant materials within a chunk (see Figure~\ref{fig:fig11}). Specifically, we use cross-correlation coefficient matching~\cite{Bradski:2000} that compares each texture tile against the Global AM representation, producing a match score map that highlights the most relevant materials. This processed information is then incorporated as additional context during training, guiding the network to prioritize the most important materials. This ensures coherent and color-stable generation, preventing the model from producing inconsistent or unnatural textures (see Figure~\ref{fig:fig12}). 
\end{description}

\section{Multi-Chunk Generation}
    \label{sec:multichunk}

While the \emph{BrushGAN} and \emph{BrushCLDM} models yield high-quality single-chunk generation, a major challenge arises when extending the generation process to multiple adjacent chunks. A naive approach, where each chunk is generated independently, leads to visible seams or stitching artifacts at chunk boundaries (see Figure~\ref{fig:fig19}(a)). These discontinuities disrupt the visual coherence of the generated game maps, necessitating a seamless multi-chunk generation strategy.

To address this, we introduce a context-aware stitching mechanism that ensures smooth transitions between neighboring chunks while preserving the structural and stylistic consistency of the generated textures.

\subsection{Multi-Chunk Generation Algorithm}
    \label{sec:mcgen}

The proposed algorithm for multi-chunk generation follows a structured pipeline to maintain texture continuity across chunks. The core steps of our approach are as follows:

\begin{enumerate}[leftmargin=*]
    \item \textbf{Sequential Content Generation:} The initial step involves generating individual chunks one-by-one using either the \emph{BrushGAN} or \emph{BrushCLDM} model, treating each chunk as an isolated generation task.
    \item \textbf{Adjacency Identification:} Afterwards, adjacent chunk pairs within the user-defined brush mask are systematically identified by evaluating their spatial positions: 
        \begin{enumerate}[leftmargin=3em]
            \item Horizontal adjacency: If $|x_1 - x_2| = 1$ and $|y_1 - y_2| = 0$,
            \item Vertical adjacency: If $|y_1 - y_2| = 1$ and $|x_1 - x_2| = 0$,
        \end{enumerate}
        where $\left(x_1, y_1\right)$ and $\left(x_2, y_2\right)$ are the spatial coordinates of the considered chunks.
    \item \textbf{Mutual Positioning Analysis:} Next, the relative direction of adjacent chunks (left, right, above, or below) is established to define boundary alignment.
    \item \textbf{Mask Intersection Check:} Overlapping of user brush masks between adjacent chunks is then analyzed in a pixel-wise manner to identify the shared regions that require seamless generation.
    \item \textbf{Stitching Mechanism:} Chunks satisfying the adjacency and mask intersection criteria are then seamlessly blended using our context-aware stitching approach (cf. Section~\ref{sec:mcstitching}) but only for the intersected materials between the two chunks.
    \item \textbf{Handling Non-Intersecting Tile Materials:} If certain tile materials are exclusive to one chunk and absent in the adjacent chunk, we apply progressive Gaussian material smoothing (see Section~\ref{sec:mcsmoothing}). 
\end{enumerate}

\subsection{Context-Aware Stitching}
    \label{sec:mcstitching}

\begin{figure}[b]
    \centering
    \includegraphics[width=1.0\linewidth]{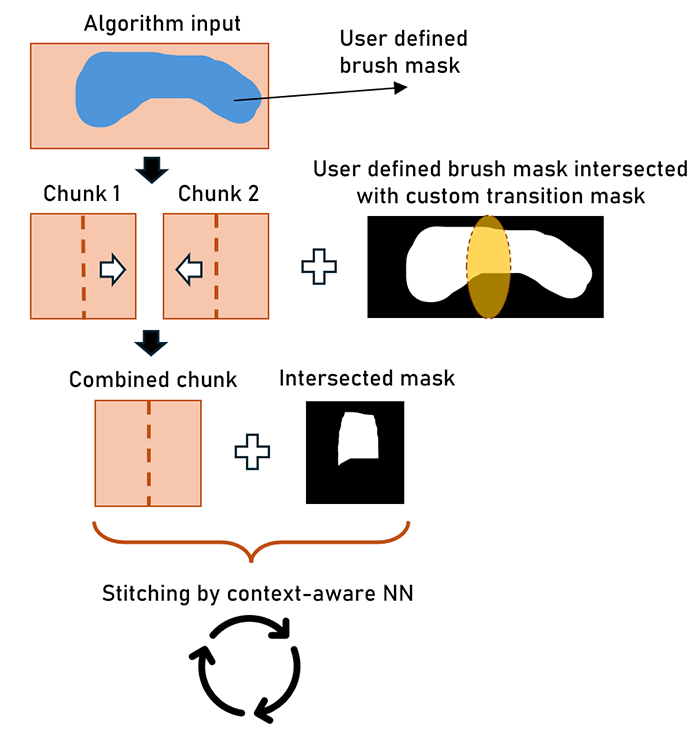}
    \caption{Context-aware stitching scheme: A custom transition mask (predefined ellipse) is generated along the boundary of two adjacent chunks and intersected with the user-defined brush mask. Afterwards, half of each chunk is extracted and stacked to form the input for the generative model (\emph{BrushGAN} or \emph{BrushCLDM}), which, in turn, synthesizes a seamless transition texture to eliminate visible stitching artifacts.}
    \label{fig:fig16}
\end{figure}

\begin{figure}
    \centering
    \includegraphics[width=1.0\linewidth]{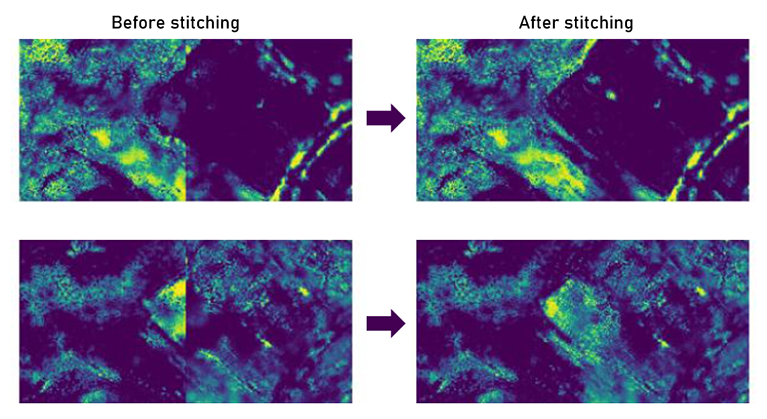}
    \hspace{0.25\linewidth}(a)\hspace{0.4\linewidth}(b)\hspace{0.25\linewidth}\\
    \caption{Comparison of chunk boundaries before and after stitching: (a) Before stitching the adjacent tiles show visible stitching artifacts due to independent chunk generation, while after stitching (b) a smooth and coherent transition between chunks is obtained.}
    \label{fig:fig17}
\end{figure}

The core idea in our method lies in the stitching mechanism (depicted in Figure~\ref{fig:fig16}) that eliminates seams at chunk boundaries. This process begins with the generation of a custom transition mask, shaped as a predefined ellipse, along the border of two adjacent chunks which is intersected with the user-defined brush mask. The elliptical shape was chosen as it provides a minimal form that fully covers the chunk intersection while reducing the generated area, thereby preserving as much of the original content as possible. Afterwards, we extract half of one chunk and half of the adjacent chunk. The extracted patches are then stacked together and fed as input into the generative model (\emph{BrushGAN} or \emph{BrushCLDM}), which synthesizes an interpolated transition texture that blends the two chunks seamlessly (see Figure~\ref{fig:fig17}).

This contextual blending step ensures that the generated content not only fills the missing region but also adapts to the structural and material properties of both chunks, mitigating any visible seams (especially on objects like roads, fields, etc.).

\subsection{Progressive Gaussian Material Smoothing}
    \label{sec:mcsmoothing}

\begin{figure}
    \centering
    \includegraphics[width=1.0\linewidth]{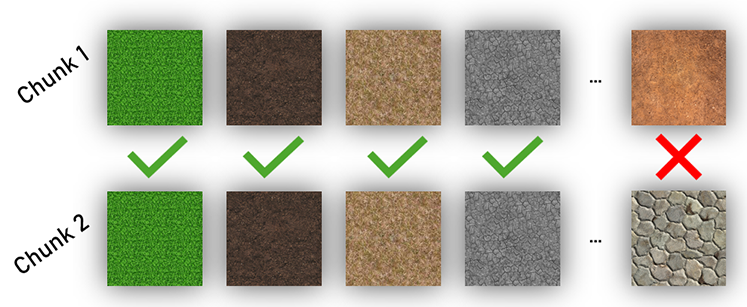}
    \caption{An example of partially intersecting material sets: The top and bottom sets represent materials for two adjacent chunks. When these sets are not identical, non-shared materials lead to visible stitching artifacts (as seen in Figure~\protect\ref{fig:fig19}(a)) at the border in complete mask mode generation.}
    \label{fig:fig18}
\end{figure}

An additional challenge arises when adjacent chunks contain non-fully overlapping tile materials, leading to potential discontinuities in material properties at chunk boundaries (see Figure~\ref{fig:fig18}). To address this issue, we use a progressive Gaussian material smoothing strategy.

Initially, we identify tile materials that are positioned along chunk borders and if a material exists only in one chunk and not in its neighbor, it is flagged as a potential stitching artifact source.

After that, a Gaussian blur kernel is progressively applied to the identified border materials. The material intensity is gradually reduced to zero at the chunk boundary, ensuring a smooth transition between materials. Next, the material strength is restored progressively towards the center of the chunk, ensuring that textures retain their original visual characteristics without abrupt cuts.

The progressive Gaussian decay function applied to the material distribution ensures a continuous, high-fidelity transition between chunks by eliminating harsh seams while at the same time preserving the original material structures.

\begin{figure*}
    \centering
    \begin{minipage}[t]{0.32\linewidth}
        \centering
        \includegraphics[width=1.0\linewidth]{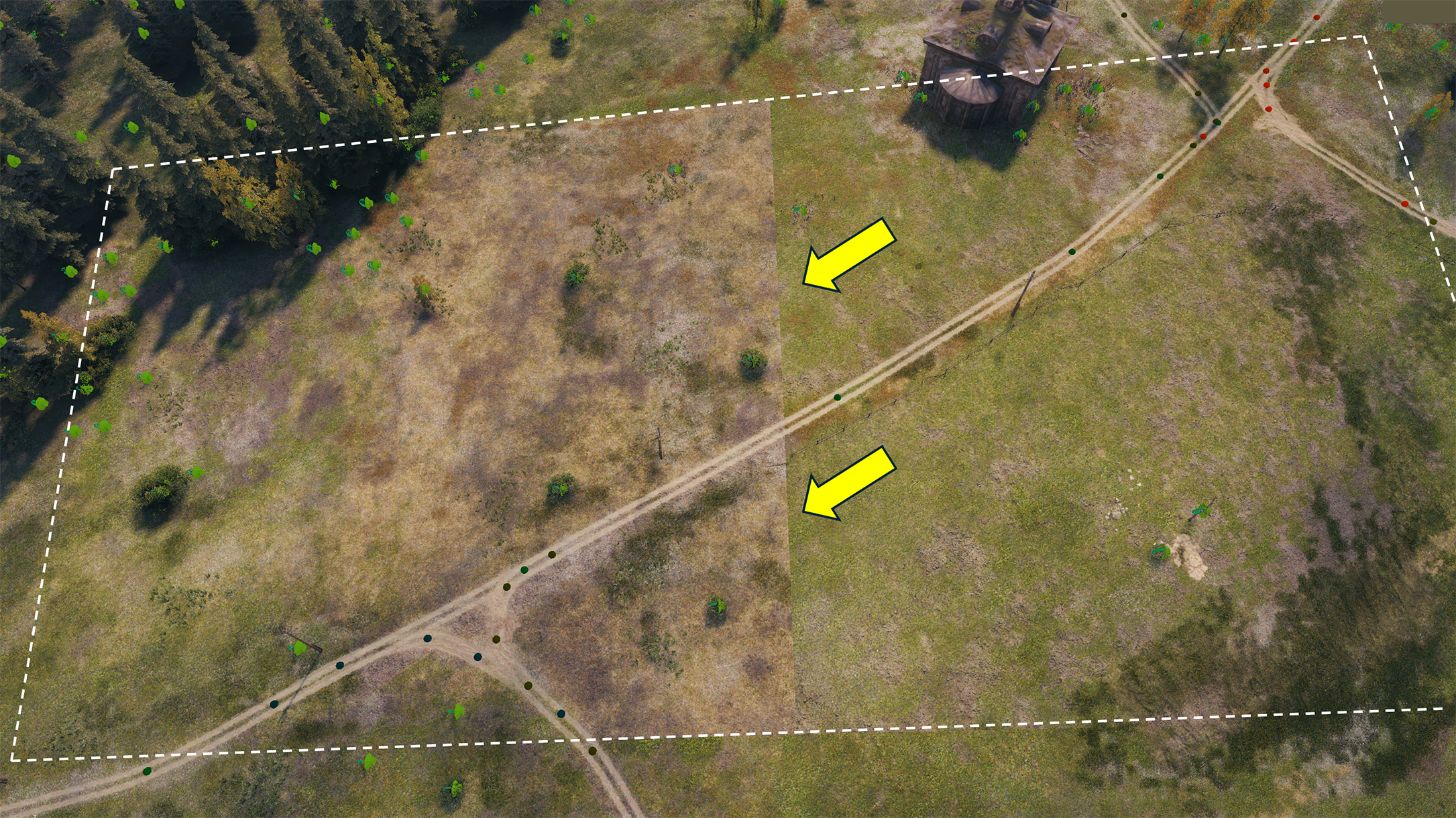}\\
        (a)
    \end{minipage}
    \hfill
    \begin{minipage}[t]{0.32\linewidth}
        \centering
        \includegraphics[width=1.0\linewidth]{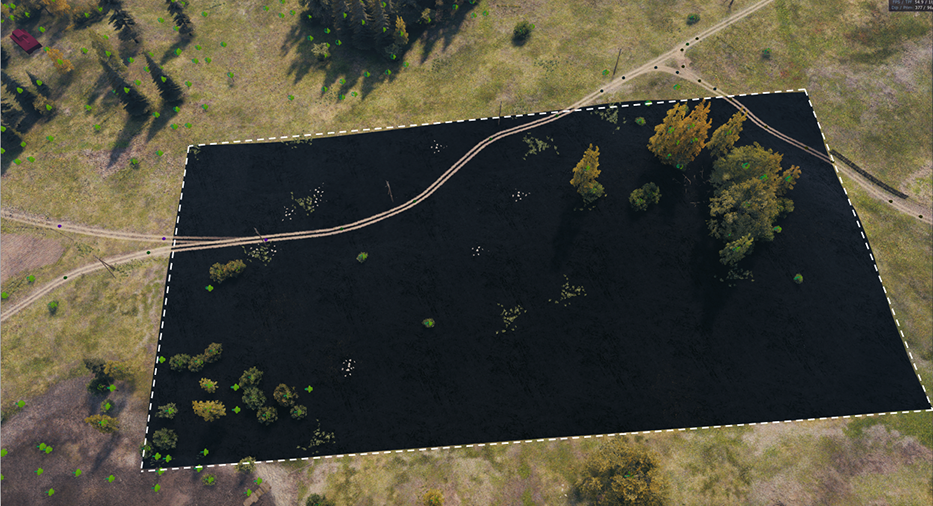}\\
        (b)
    \end{minipage}
    \hfill
    \begin{minipage}[t]{0.32\linewidth}
        \centering
        \includegraphics[width=1.0\linewidth]{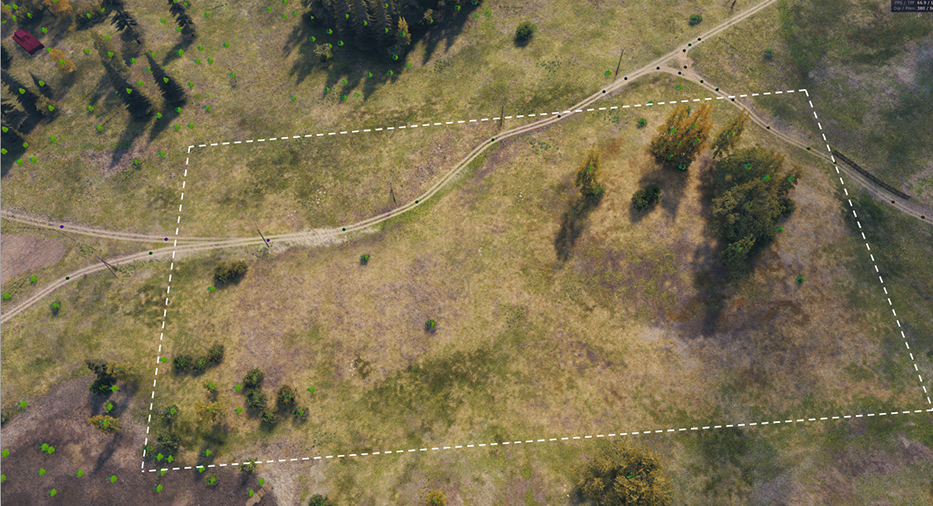}\\
        (c)
    \end{minipage}
    \caption{Seamless Multi-Chunk Generation: (a) An example of visible stitch for independent chunk generation in complete mask mode. (b) Input, where black regions indicate masked areas. (c) Generated results in multi-chunk mode, demonstrating that our seamless generation technology effectively eliminates visible stitching between adjacent chunks and their outer neighbors.}
    \label{fig:fig19}
\end{figure*}

Finally, our combined approach effectively eliminates stitching artifacts, ensuring a seamless and visually coherent transition between adjacent chunks, as demonstrated in Figure~\ref{fig:fig19}(c)).

\section{Evaluation}
    \label{sec:evaluation}

\begin{table*}
    \centering
    \begin{tabularx}{0.95\linewidth}{X ccc@{\hskip20pt} ccc@{\hskip20pt} ccc}
        \toprule
        & \multicolumn{3}{c}{\textbf{medium mask mode}} & \multicolumn{3}{c}{\textbf{hard mask mode}} & \multicolumn{3}{c}{\textbf{complete mask mode}}\\
        \textbf{Model}  & \textbf{FID (↓)}   & \textbf{LPIPS (↓)} & \textbf{SSIM (↑)}  & \textbf{FID (↓)}   & \textbf{LPIPS (↓)} & \textbf{SSIM (↑)} & \textbf{FID (↓)}   & \textbf{LPIPS (↓)} & \textbf{SSIM (↑)}\\
        \midrule
        \emph{BrushGAN}     & \textbf{0.18}      & \textbf{0.11}      & \textbf{0.90}  & \textbf{1.47}	    & \textbf{0.30}	& 0.65 & \textbf{10.04}	& \textbf{0.57}	& 0.35   \\
        \emph{BrushCLDM}    & 0.90      & \textbf{0.11}      & 0.89  & 8.14	    & 0.32	& \textbf{0.67} & 46.37	& 0.61	& 0.41   \\
        \emph{SN-PatchGAN}              & 1.30      & 0.15      & 0.83  & 11.5	    & 0.38	& 0.61 & 116.71	& 0.67	& 0.29   \\
        \emph{CLDM}                & 0.94      & \textbf{0.11}      & 0.89  & 10.52	    & 0.34	& 0.64 & 149.34	& 0.64	& \textbf{0.43}   \\
        \emph{SPADE}               & 7.82      & 0.23      & 0.73  & 80.10	    & 0.60	& 0.43 & 136.87	& 0.71	& 0.26   \\
        \emph{Palette}             & 8.43      & 0.28      & 0.69  & 101.32    & 0.63	& 0.41 & 169.90	& 0.74	& 0.25   \\
        \bottomrule
    \end{tabularx}
    \caption{Performance in medium mask mode (<30\% coverage), hard mask mode (30-99\% coverage), and complete mask mode (pure generation, 100\% coverage). ↓ = smaller values are better, ↑ = larger values are better, best scores are written in bold-face.}
    \label{table:results}
\end{table*}

In the following, we evaluate our proposed approaches with respect to quality and time performance.

\subsection{Quality performance}
    \label{sec:evalquality}

To assess the effectiveness of our proposed generative approaches, we conducted an evaluation of our two models -- \emph{BrushGAN} and \emph{BrushCLDM}. Additionally, we compared our approaches to adapted versions of \emph{SPADE} and \emph{Palette}, ensuring they align with our specific generation scenario (8 input channels), as well as to the original \emph{SN-PatchGAN} and \emph{LDM} with \emph{ControlNet} (\emph{CLDM}).

Our evaluation is conducted using the test dataset described in Section~\ref{sec:dataset}, which consists of chunks with different material sets, covering all important map cases: urban, winter, natural. It allows us to systematically analyze model performance under different inpainting and generation scenarios with diverse input conditions.

Our evaluation considers three metrics:

\begin{itemize}[leftmargin=*]
    \item Frechet Inception Distance (FID)~\cite{Heusel:2017} evaluates realism by measuring the distribution distance between generated and real textures.
    \item Learned Perceptual Image Patch Similarity (LPIPS)~\cite{Zhang:2018} measures perceptual similarity to human vision.
    \item Structural Similarity Index (SSIM)~\cite{Wang:2004} assesses structural consistency with the reference textures.
\end{itemize}

\noindent We examine performance across three different masking modes:

\begin{itemize}[leftmargin=*]
    \item Medium (<30\%) –- Small missing regions with significant neighboring context.
    \item Hard (30-99\%) –- Challenging cases with partially missing structure.
    \item Complete (pure generation, 100\%) –- No neighboring pixel information, requiring full texture synthesis.
\end{itemize}

\noindent The results for medium mask mode (see Table~\ref{table:results}) indicate that the \emph{BrushGAN} model achieves the highest performance in terms of perceptual quality, followed closely by \emph{BrushCLDM}. Notably, while \emph{CLDM} and \emph{BrushCLDM} trail behind, its performance remains within a comparable range to the \emph{BrushGAN} approach. Conversely, \emph{Palette}, \emph{SPADE}, and the original implementation of \emph{SN-PatchGAN} exhibit substantially inferior performance, supposingly due to their lack of context adaptation mechanisms, making them less effective at reconstructing missing content within the masked regions. 

In the hard mask mode (see Table~\ref{table:results}), where the masked region size increases and available contextual information decreases, the \emph{BrushGAN} model remains the best performer, followed by \emph{BrushCLDM}. The context-aware nature of both models allows them to infer missing structures more effectively than the other investigated models, leading to substantially improved perceptual and structural consistency. The remaining models -- \emph{Palette}, \emph{SPADE}, the original \emph{CLDM}, and \emph{SN-PatchGAN} -- exhibit significantly lower performance, primarily due to their inability to integrate contextual cues effectively in regions with missing information. 

\begin{figure*}
    \centering
    \includegraphics[width=0.8\linewidth]{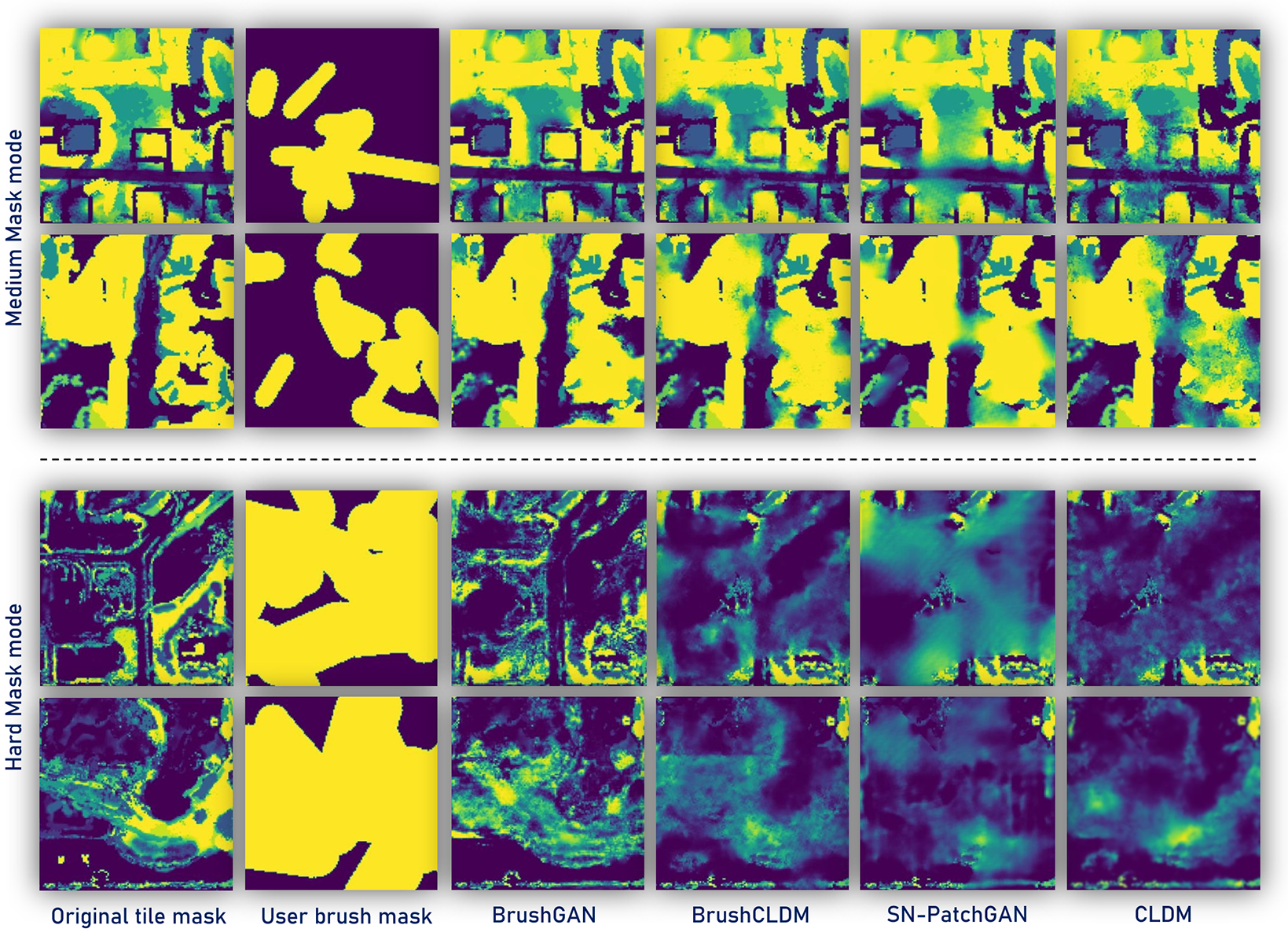}
    \caption{Comparison of \emph{BrushGAN} and \emph{BrushCLDM} in medium and hard mask modes: The first two rows present results for the medium mask mode, while the third and fourth rows correspond to the hard mask mode. We observe that \emph{BrushGAN} produces sharper and more detailed outputs while preserving image context, whereas \emph{BrushCLDM} generates smoother patterns but still follows contextual cues. In contrast, \emph{SN-PatchGAN} and \emph{CLDM} yield noticeably blurrier results and exhibit difficulties in maintaining contextual consistency.}
    \label{fig:fig13}
\end{figure*}

For the complete mask mode (see Table~\ref{table:results}), where no neighboring pixel information is available, the \emph{BrushGAN} model again outperforms all alternatives, followed by \emph{BrushCLDM}. As in the hard mask case, the context-aware mechanisms in these models allow for higher semantic consistency and texture reconstruction. The other models (\emph{Palette}, \emph{SPADE}, \emph{CLDM}, and \emph{SN-PatchGAN}) show substantial degradation, reinforcing the importance of incorporating context-adaptive strategies for handling empty regions effectively.

Based on our observations, we believe that FID aligns more closely with our subjective aesthetic perception of image quality compared to other metrics, as it effectively captures both feature-level similarity and perceptual coherence in generated content.

Our evaluation demonstrates that the \emph{BrushGAN} model consistently outperforms all evaluated alternatives across all mask scenarios, making it the most effective approach for both inpainting and full-content generation. This can largely be attributed to its ability to incorporate contextual information, enabling it to generate highly realistic textures even in hard and complete mask cases, where neighboring information is limited or entirely absent.

The \emph{BrushCLDM} model follows closely, performing well in all cases, though slightly inferior to \emph{BrushGAN} due to its weaker fine-detail preservation. Nevertheless, \emph{BrushCLDM} retains its ability to leverage contextual features, making it a strong choice for scenarios requiring structure-aware generation.

All other models, namely the original \emph{SN-PatchGAN}, \emph{CLDM}, \emph{Palette}, and \emph{SPADE} perform notably worse across all conditions. This is primarily due to two key limitations: (1) these models are not explicitly designed to handle complex data distributions as seen in our test dataset, and (2) they lack robust context-awareness mechanisms, which is crucial for generating high-quality textures in the challenging hard and complete mask cases. As a result, these models exhibit poor texture consistency and structural coherence, particularly when faced with large missing regions, where their outputs deviate strongly from artist-intended style and structural integrity.

Figure~\ref{fig:fig13} presents a qualitative comparison of the top-performing models, \emph{BrushGAN} and \emph{BrushCLDM}, in medium and hard mask modes. The examples highlight the distinct characteristics of the two approaches in terms of detail restoration. \emph{BrushGAN} demonstrates a stronger capability in reconstructing high-frequency textures, whereas \emph{BrushCLDM} prioritizes the preservation of low-frequency structural details.

\begin{figure*}
    \centering
    \begin{minipage}[t]{0.27\linewidth}
        \centering
        \includegraphics[width=1.0\linewidth]{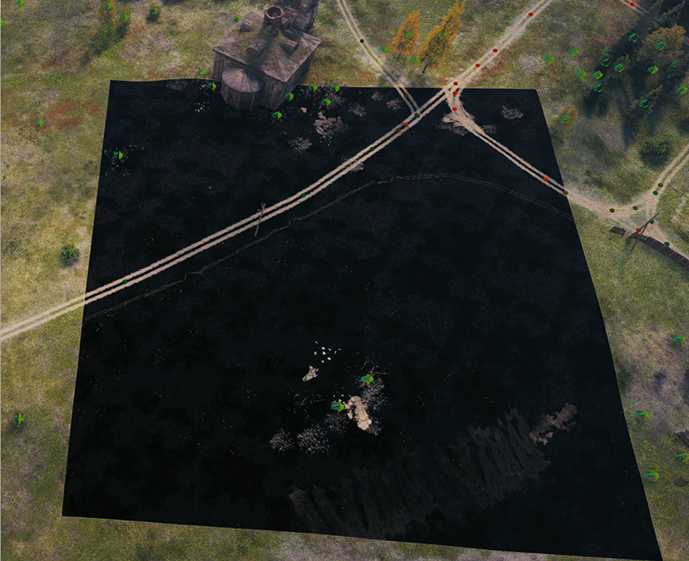}\\
        (a)
    \end{minipage}
    \hspace{0.5cm}
    \begin{minipage}[t]{0.27\linewidth}
        \centering
        \includegraphics[width=1.0\linewidth]{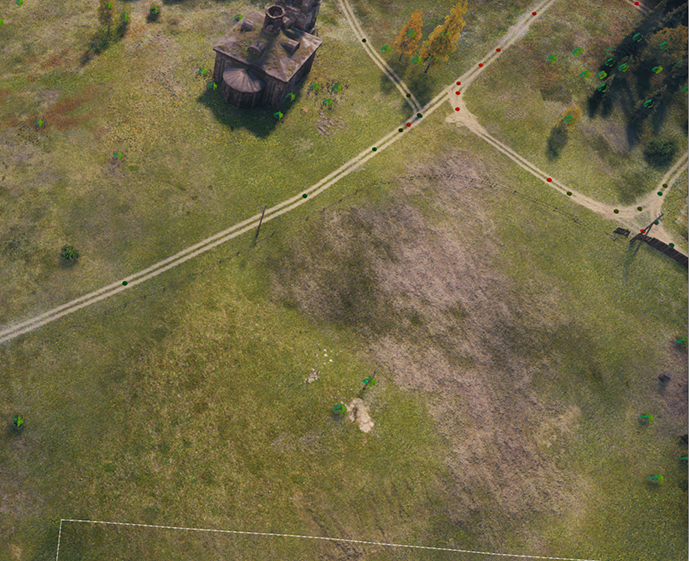}\\
        (b)
    \end{minipage}
    \hspace{0.5cm}
    \begin{minipage}[t]{0.27\linewidth}
        \centering
        \includegraphics[width=1.0\linewidth]{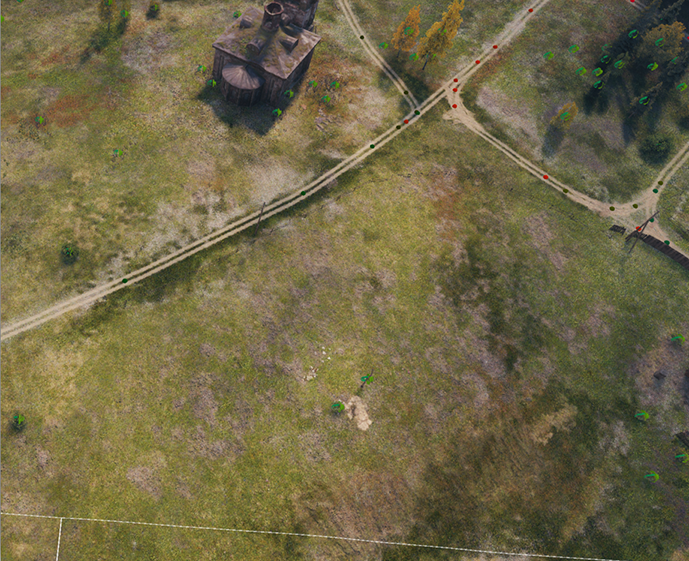}\\
        (c)
    \end{minipage}
    \caption{Comparison of \emph{BrushGAN} and \emph{BrushCLDM} in complete mask mode: (a) input with missing textures (black area), (b) \emph{BrushCLDM} generated result, and (c) \emph{BrushGAN} generated result.}
    \label{fig:fig14}
\end{figure*}

In Figure~\ref{fig:fig13} we demonstrated the restoration of tile masks. However, in the complete mask mode, showcasing the final gaming map results provides a more visually compelling and comprehensive representation of the real impact of our \emph{BrushGAN} and \emph{BrushCLDM} models on the overall map quality and seamless integration. This can be observed in Figure~\ref{fig:fig14}.

Overall, our findings highlight the necessity of context-adaptive architectures in addressing challenging generative tasks for gaming map development.

Note that we focused on the evaluation of single-chunk generation above, since assessing multi-chunk generation quantitatively is challenging. This is because our multi-chunk approach relies on sequential single-chunk evaluations combined with adaptive boundary blending, its performance inherently incorporates the effectiveness of the single-chunk models. Furthermore, direct comparisons with existing methods are non-trivial, as standard evaluation metrics (e.g., FID, LPIPS, SSIM) are designed for independent sample assessment and do not account for spatial consistency across multiple generated chunks. 

\subsection{Time Performance}
    \label{sec:evaltime}

Efficiency is a crucial factor in practical applications of generative models. The \emph{BrushGAN} model achieves a processing time of about 2 seconds per chunk on an A100 GPU, while \emph{BrushCLDM} requires roughly 6 seconds (averaged across 15 independent inference runs). Despite this difference, both models remain highly viable for real-world use, as artists typically spend hours designing a single chunk from scratch. This demonstrates that our approach contributes to accelerating the creative process while maintaining high-quality texture synthesis.

\section{Conclusions}
    \label{sec:conclusions}

In this paper, we introduced \emph{BrushGAN} and \emph{BrushCLDM}, two novel AI-driven generative models for seamless gaming map generation to address the challenges of both single-chunk and multi-chunk generation. Through context-aware modeling and adaptive blending techniques, our approach ensures high-quality texture synthesis while eliminating common artifacts such as stitching seams and material discontinuities. To enable multi-chunk generation, we proposed a multi-component blending technique that ensures smooth transitions between adjacent chunks.

We demonstrated that our proposed models outperform existing methods in both inpainting and pure generation tasks, substantially improving structural coherence and texture fidelity. With an average processing time of 2 to 6 seconds per chunk -- depending on method -- and considering the time-consuming process of traditional manual game map design, our AI-assisted approach contributes to accelerating content creation while maintaining artist-level quality.

In conclusion, we believe that our approach successfully bridges the gap between AI-based texture generation and artist-driven design, offering a scalable solution for game development and virtual environments. Future work will focus on enhancing our approach by incorporating more complex contextual information (wetness maps, decals and debris types and positions, etc.), enabling the generation of entire maps from scratch while maintaining high-resolution detail and artistic quality.

\begin{acks}
    We would like to thank Wargaming for providing a high-quality dataset for experimentation and evaluation, as well as for granting access to computational resources that facilitates this research.
\end{acks}

\balance
\bibliographystyle{ACM-Reference-Format}
\bibliography{references}

\end{document}